%% file: main.tex
\documentclass{article}

\PassOptionsToPackage{numbers, compress}{natbib}
\usepackage[final]{neurips_2023}

\usepackage[numbers,compress]{natbib}

\usepackage[hyperfootnotes=false]{hyperref}
\hypersetup{
  colorlinks,
  citecolor=green,
  linkcolor=violet,
  urlcolor=blue
}

\usepackage[utf8]{inputenc} % allow utf-8 input
\usepackage[T1]{fontenc}    % use 8-bit T1 fonts
\usepackage{hyperref}       % hyperlinks
\usepackage{url}            % simple URL typesetting
\usepackage{booktabs}       % professional-quality tables
\usepackage{amsfonts}       % blackboard math symbols
\usepackage{nicefrac}       % compact symbols for 1/2, etc.
\usepackage{microtype}      % microtypography
\usepackage{xcolor}         % colors

\usepackage{graphicx}
\usepackage{subfigure}
\usepackage{kotex}
\usepackage{duckuments}
\usepackage{amsmath}
\usepackage{amssymb}
\usepackage{mathtools}
\usepackage{amsthm}
\usepackage[capitalize,noabbrev]{cleveref}
\usepackage[textsize=tiny]{todonotes}

\usepackage{wrapfig}
\usepackage{lipsum}
\usepackage[export]{adjustbox}

\usepackage{listings}
\usepackage{xcolor}

\definecolor{codegreen}{rgb}{0,0.6,0}
\definecolor{codegray}{rgb}{0.5,0.5,0.5}
\definecolor{codepurple}{rgb}{0.58,0,0.82}
\definecolor{backcolour}{rgb}{0.95,0.95,0.92}

\lstdefinestyle{mystyle}{
    backgroundcolor=\color{backcolour},   
    commentstyle=\color{codegreen},
    keywordstyle=\color{magenta},
    numberstyle=\tiny\color{codegray},
    stringstyle=\color{codepurple},
    basicstyle=\ttfamily\footnotesize,
    breakatwhitespace=false,         
    breaklines=true,                 
    captionpos=b,                    
    keepspaces=true,                 
    numbers=left,                    
    numbersep=5pt,                  
    showspaces=false,                
    showstringspaces=false,
    showtabs=false,                  
    tabsize=2
}

\theoremstyle{plain}

\theoremstyle{definition}

\theoremstyle{remark}

\usepackage{algorithmic}% http://ctan.org/pkg/algorithms
\usepackage{algorithm}% http://ctan.org/pkg/algorithms

\input{macros.tex}

\title{
    Understanding the Latent Space of Diffusion Models through the Lens of Riemannian Geometry
}

\author{
Yong-Hyun Park$^{*1}$,~
Mingi Kwon$^{*2}$,~
Jaewoong Choi$^{3}$,~
Junghyo Jo$^{\dagger1}$,~
Youngjung Uh$^{\dagger2}$,
\vspace{+0.3em}
\\
\normalsize$^1$Seoul National University~~
$^2$Yonsei University~~
$^3$Korea Institute for Advanced Study\\
}

\begin{document}

\maketitle

% \blfootnote{$^*$Equal Contribution\hspace{3mm} $^dagger$ Corresponding authors}

\begin{abstract}
Despite the success of diffusion models (DMs), we still lack a thorough understanding of their latent space. To understand the latent space $\mathbf{x}_t \in \mathcal{X}$, we analyze them from a geometrical perspective. Our approach involves deriving the local latent basis within $\mathcal{X}$ by leveraging the pullback metric associated with their encoding feature maps. Remarkably, our discovered local latent basis enables image editing capabilities by moving $\mathbf{x}_t$, the latent space of DMs, along the basis vector at specific timesteps. We further analyze how the geometric structure of DMs evolves over diffusion timesteps and differs across different text conditions. This confirms the known phenomenon of coarse-to-fine generation, as well as reveals novel insights such as the discrepancy between $\mathbf{x}_t$ across timesteps, the effect of dataset complexity, and the time-varying influence of text prompts. To the best of our knowledge, this paper is the first to present image editing through $\mathbf{x}$-space traversal, editing only once at specific timestep $t$ without any additional training, and providing thorough analyses of the latent structure of DMs.
The code to reproduce our experiments can be found at {\color{cyan}\url{https://github.com/enkeejunior1/Diffusion-Pullback}}.
\end{abstract}

\nnfootnote{$^*$Equal Contribution\hspace{3mm} $^\dagger$ Corresponding authors}
% \vspace{-2.5em}

\input{1intro}
\input{2relatedwork}
\input{3method}

\input{4experiment}
\input{5conclusion}

\bibliography{reference_papers}
\bibliographystyle{plainnat}

\input{6appendix}

\end{document}

%% file: macros.tex
\usepackage{amsmath,amsfonts,bm}

\newcommand\nnfootnote[1]{%
  \begin{NoHyper}
  \renewcommand\thefootnote{}\footnote{#1}%
  \addtocounter{footnote}{-1}%
  \end{NoHyper}
}

\newcommand{\fref}[1]{Figure~\ref{#1}}

\newcommand{\eref}[1]{Eq.~(\ref{#1})}
\newcommand{\tref}[1]{Table~\ref{#1}}
\newcommand{\sref}[1]{$\S$~\ref{#1}}
\newcommand{\aref}[1]{Appendix \ref{#1}}

\newcommand{\ehspace}{$\mathcal{H}$}
\newcommand{\exspace}{$\mathcal{X}$}

\newcommand\uh[1]{\textcolor{black}{#1}}
\newcommand\mingi[1]{\textcolor{black}{#1}}
\newcommand\modify[1]{\textcolor{black}{#1}}
\newcommand\yh[1]{\textcolor{black}{#1}}
\newcommand\jo[1]{\textcolor{black}{#1}}

\def\jacx{{J}_{\mathbf{x}}}
\def\tanxspace{\mathcal{T}_\mathbf{x}}
\def\tanhspace{\mathcal{T}_\mathbf{h}}

\def\vx{\mathbf{x}}
\def\vh{\mathbf{h}}

\def\vv{\mathbf{v}}
\def\vu{\mathbf{u}}
\def\dx{\mathbf{v}}
\def\dh{\mathbf{u}}
\def\predictedx0{x_0}
\newcommand*{\tran}{^{\mkern-1.5mu\mathsf{T}}}

%% file: 1intro.tex
\section{Introduction}
The diffusion models (DMs) are powerful generative models that have demonstrated impressive performance \cite{ho2020denoising, song2020denoising, song2020score, dhariwal2021diffusion, nichol2021improved}. 
DMs have shown remarkable applications, including text-to-image synthesis \cite{ramesh2022hierarchical, rombach2022high, balaji2022ediffi, nichol2021glide}, inverse problems \cite{chung2022improving, lugmayr2022repaint}, and image editing \cite{hertz2022prompt, tumanyan2022plug, parmar2023zero, mokady2022null}.

% Diffusion models (DMs) are highly powerful generative models that have shown great performance \cite{ho2020denoising,song2020denoising,song2020score,dhariwal2021diffusion,nichol2021improved}.
% \yh{Instead of understanding the semantic structure of latent space itself,} 
% To control the generative process,
% existing methods have introduced conditional DMs, especially for text-to-image synthesis \cite{ramesh2022hierarchical,rombach2022high,balaji2022ediffi,nichol2021glide}, or mixing the latent variables $\vx_t$ of different sampling processes \cite{choi2021ilvr,meng2021sdedit,avrahami2022blended,liew2022magicmix,kawar2022imagic,avrahami2022spatext}.

Despite their achievements, the research community lacks a comprehensive understanding of the latent space of DMs and its influence on the generated results. So far, the completely diffused images are considered as latent variables but it does not have useful properties for controlling the results. For example, traversing along a direction from a latent produces weird changes in the results.
Fortunately, \citet{kwon2022diffusion} consider the intermediate feature space of the diffusion kernel, referred to as \ehspace{}, as a semantic latent space and show its usefulness on controlling generated images. In the similar sense, some works investigate the feature maps of the self-attention or cross-attention operations for controlling the results \cite{hertz2022prompt, tumanyan2022plug, parmar2023zero}, improving sample quality \cite{chefer2023attend}, or downstream tasks such as semantic segmentation \cite{ma2023diffusionseg, xu2023open}.
% They pair this with an asymmetric sampling process and
% They demonstrated that \ehspace{} possesses a linear semantic structure.
% Following the work of \citet{kwon2022diffusion}, there have been several analyses of the DM's feature space. These investigations explore the attention map and cross-attention map, which have been used for image editing \cite{hertz2022prompt, tumanyan2022plug, parmar2023zero} or improving sample quality \cite{chefer2023attend}. Moreover, these findings can be applied to downstream tasks such as segmentation using the feature space \cite{ma2023diffusionseg, xu2023open}.

% However, they rely solely on a proxy, $\mathbf{h}$, and do not directly deal with the latent variables $\vx_t$. 
% Furthermore, due to Kwon et al.'s modification of the internal feature of UNet, particularly where skip connections exist, there is no guarantee that significant changes will result in output quality. It is worth noting that the latent space \exspace{} is a sparse manifold, which makes meaningful editing within it challenging. Nevertheless, since \ehspace{} is clearly the codomain of \exspace{}, and given the fact that a meaningful semantic direction can be found in \ehspace{}, we can also define a semantic subspace in \exspace{}.

Still, the structure of the space $\mathcal{X}_t$ where latent variables $\{\vx_t\}$ live remains unexplored despite its crucial role in understanding DMs. It is especially challenging because 1) the model is trained to estimate the forward noise which does not depend on the input, as opposed to other typical supervisions such as classification or similarity, and 2) there are lots of latent variables over multiple recursive timesteps.
% Although the feature space has been thoroughly explored, the latent space $\vx{}_t \in \mathcal{X}$ remains unexplored despite its crucial role in understanding DMs.
% This is due to the fact that the latent variable of DM, $\vx{}_t$, consists primarily of noised images.
% Therefore, it is not reasonable to expect any significant semantic structure in the latent space \exspace{}.
% \yh{Another reason is the the characteristic iterative process of the DMs which involves a sequence of noisy images and subtle noises, i.e., the embeddings are not directly connected to the final images.}
In this paper, we aim to analyze \exspace{} in conjunction with its corresponding representation \ehspace{}, by incorporating a local geometry to \exspace{} using the concept of a {\it pullback metric} in Riemannian geometry. 
% Our paper contributes:

\begin{figure}[!t]
    \centering
    \includegraphics[width=1\linewidth]{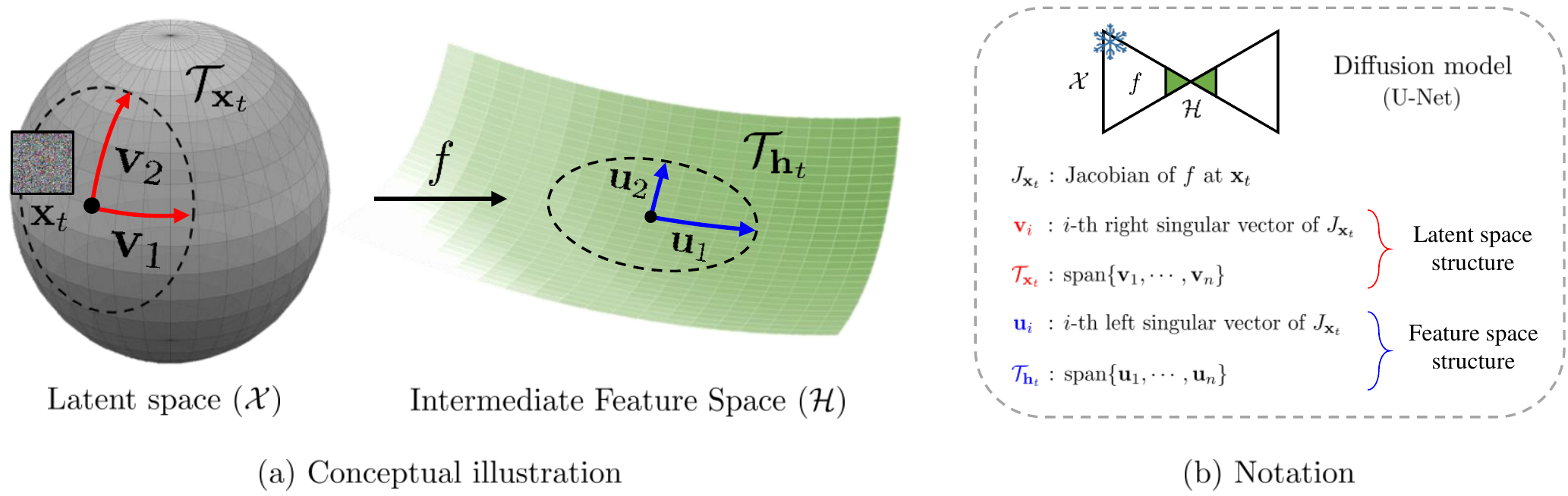}
    \caption{\textbf{Conceptual illustration of local geometric structure.} 
    (a) The local basis $\{\mathbf{v}_1, \mathbf{v}_2, \cdots \}$ of the local latent subspace $\mathcal{T}_{{\mathbf{x}}_t}$ within the latent space $\mathcal{X}$ is interlinked with the local basis $\{\mathbf{u}_1, \mathbf{u}_2, \cdots\}$ of the local tangent space $\mathcal{T}_{{\mathbf{h}}_t}$ in the feature space $\mathcal{H}$.
    % (b) The derivation of these local bases is facilitated through the singular value decomposition (SVD) of the Jacobian originating from the U-Net responsible for encoding the feature map $f$, which connects $\mathcal{X}$ and $\mathcal{H}$.
    (b) The derivation of these local bases is facilitated through the singular value decomposition (SVD) of the Jacobian, which emanates from the U-Net responsible for encoding the feature map \(f\), linking \(\mathcal{X}\) and \(\mathcal{H}\).
    }
    \vspace{-1em}
    \label{fig:teaser}
\end{figure}

% \begin{figure}[!t]
%     \centering
%     % \includegraphics[width=1\linewidth]{figure/teaser_new.pdf}
%     \includegraphics[width=1\linewidth]{figure/teaser_rebuttal.pdf}
%     \caption{\textbf{Conceptual illustration of our method.} 
%     % We employ an unsupervised approach to discover latent basis vectors in the latent space $\vx_t$, leveraging the Riemannian geometry between $\vx_t$ and $\vh_t$. Here, $\mathcal{H}$ represents the bottleneck layer, and $f$ denotes the frozen encoder of a U-Net. 
%     % The identified directions allow for the manipulation of the semantic content in the generated images. Text conditioning with prompts can be also used for targeted image editing.
%     We employ an unsupervised approach to discover the latent basis in the latent space $\mathcal{X}_t$.
%     }
%     \vspace{-1em}
%     \label{fig:teaser}
% \end{figure}

First, we discover the local latent basis for \exspace{} and the corresponding local tangent basis for \ehspace{}.
The local basis is obtained by performing singular value decomposition (SVD) of the Jacobian of the mapping from \exspace{} to \ehspace{}.
To validate the discovered local latent basis, we demonstrate that 
walking along the basis
% \yh{traversal along the basis} 
can edit real images in a semantically meaningful way. 
Furthermore, we can use the discovered local latent basis vector to edit other samples by using parallel transport, when they exhibit comparable local geometric structures.
Note that existing editing methods manipulate the self-attention map or cross-attention map over multiple timesteps \cite{hertz2022prompt, tumanyan2022plug, parmar2023zero}. On the other hand, we manipulate only $\vx_t$ once at a specific timestep.

Second, we investigate how the latent structures differ across different timesteps and samples as follows.
% Second, we investigate how the semantic structure changes over time and \jo{between samples}. %identified two trends.
The frequency domain of the local latent basis shifts from low-frequency to high-frequency along the generative process. 
% Although this was observed indirectly in a previous study by \citet{choi2022perception},
We explicitly confirm it using power spectral density analysis.
The difference between local tangent spaces of different samples becomes larger along the generative process. 
The local tangent spaces at various diffusion timesteps are similar to each other if the model is trained on aligned datasets such as CelebA-HQ or Flowers. However, this homogeneity does not occur on complex datasets such as ImageNet.

% We compared the evolution of the local tangent basis over time and identified groups of timesteps exhibiting with similar semantic structures. 
% Within each group, DMs processed similar basis, even when given different timesteps.
% TODO ImageNet 에서의 양상 확인
% gDDIM 의 denoising process 를 나누는 기준을 떠올리게 한다. 
% TODO
% 이 관찰으로부터 유사한 semantic structure 를 갖는 구간을 clustering 하고, 각 cluster 에서 redundant 하지 않은 timestep scheduling 을 제안하고 sampling quality 를 개선했다.

Finally, we examine how the prompts affect the latent structure of text-to-image DMs as follows.
Similar prompts yield similar latent structures. Specifically, we find a positive correlation between the similarity of prompts and the similarity of local tangent spaces.
The influence of text on the local tangent space becomes weaker along the generative process. %, specifically when the diffusion timestep is $t<0.7T$. $T$ is the maximum length of a diffusion process.
%\textcircled{\raisebox{-0.9pt}{1}} 
%\textcircled{\raisebox{-0.9pt}{2}} 

% \modify{With ChatGPT LOL :) 고쳐야함.ㅋㅋ}
% 우리의 연구는 지금까지 탐구되지 않았던 DM x 에 대한 
% Our research uncovers the hidden depths of the latent space \exspace{} within DMs, revealing a profound understanding and intuition that has remained elusive. This newfound knowledge not only empowers us to achieve real image editing through unsupervised methods but also offers a glimpse into the underlying principles of generative models, opening up a realm of endless possibilities and uncharted territories. % 

{
% Our work examines the geometric structure of \exspace{} and \ehspace{} via Riemannian geometry. 
% As a result, we discover the latent structure of \exspace{}, and how the geometric structure evolves along a generative process and is affected by prompts. 
% This geometrical perspective deepen our understanding of DMs.
% Our work examines the geometry of \exspace{} and \ehspace{} through Riemannian geometry. As a result, we discover the latent structure of \exspace{} and how the geometric structure evolves along a generative process and is affected by prompts. 
% Our exploration from a geometrical perspective deepens our understanding of DMs.
Our work examines the geometry of \exspace{} and \ehspace{} using Riemannian geometry. We discover the latent structure of \exspace{} and how it evolves during the generative process and is influenced by prompts. This geometric exploration deepens our understanding of DMs.
}

% In the experiments, we demonstrate that the directions found in an unsupervised manner indeed lead to semantic changes in the images. 
% We note that discovering the editing directions in the latent variables of diffusion models has not been tackled.
% \mingi{}
% Furthermore, we provide thorough quantitative and qualitative analyses on the aforementioned properties. Our method even works on stable diffusion \cite{rombach2022high}.

%% file: 2relatedwork.tex
\section{Related works}
\paragraph{Diffusion Models.}
Recent advances in DMs make great progress in the field of image synthesis and show state-of-the-art performance \cite{sohl2015deep, ho2020denoising, song2020denoising}. 
An important subject in the diffusion model is the introduction of gradient guidance, including classifier-free guidance, to control the generative process \cite{dhariwal2021diffusion,sehwag2022generating,avrahami2022blended,liu2021more,nichol2021glide,rombach2022high}. 
The work by \citet{song2020score} has facilitated the unification of DMs with score-based models using SDEs, enhancing our understanding of DMs as a reverse diffusion process.
However, the latent space is still largely unexplored, and our understanding is limited.

\paragraph{The study of latent space in GANs.} The study of latent spaces has gained significant attention in recent years. In the field of Generative Adversarial Networks (GANs), researchers have proposed various methods to manipulate the latent space to achieve the desired effect in the generated images \cite{ramesh2018spectral,patashnik2021styleclip,abdal2021styleflow, harkonen2020ganspace,shen2021closed,yuksel2021latentclr, pan2023drag}.
% \modify{For example, local latent space manipulation techniques \cite{ramesh2018spectral,patashnik2021styleclip,abdal2021styleflow} have been developed, as well as global manipulation techniques  \cite{harkonen2020ganspace,shen2021closed,yuksel2021latentclr}. 이젠 굳이 local global 나누지 말죠} 
More recently, several studies \cite{zhu2021low, choi2021not} have examined the geometrical properties of latent space in GANs and utilized these findings for image manipulations. These studies bring the advantage of better understanding the characteristics of the latent space and facilitating the analysis and utilization of GANs. In contrast, the latent space of DMs remains poorly understood, making it difficult to fully utilize their capabilities.

\paragraph{Image manipulation in DMs.}
Early works include \citet{choi2021ilvr} and \citet{meng2021sdedit} have attempted to manipulate the resulting images of DMs by replacing latent variables, allowing the generation of desired random images. 
However, due to the lack of semantics in the latent variables of DMs, current approaches have critical problems with semantic image editing.
Alternative approaches have explored the potential of using the feature space within the U-Net for semantic image manipulation. For example, \citet{kwon2022diffusion} have shown that the bottleneck of the U-Net, \ehspace{}, can be used as a semantic latent space. \modify{Specifically, they used CLIP \cite{radford2021learning} to identify directions within $\mathcal{H}$ that facilitate genuine image editing.} \citet{baranchuk2021label} and \citet{tumanyan2022plug} use the feature map of the U-Net for semantic segmentation and maintaining the structure of generated images. 
% Unlike previous works, our editing method directly traverses the latent variable along the latent basis.
Unlike previous works, our editing method \modify{finds the editing direction without supervision}, and directly traverses the latent variable along the latent basis.

\paragraph{Riemannain Geometry.} Some studies have applied Riemannian geometry to analyze the latent spaces of deep generative models, such as Variational Autoencoders (VAEs) and GANs \cite{arvanitidis2017latent, shao2018riemannian, chen2018metrics, arvanitidis2020geometrically, lee2023explicit, pmlr-v162-lee22d, yonghyeon2021regularized}. \citet{shao2018riemannian} proposed a pullback metric on the latent space from image space Euclidean metric to analyze the latent space's geometry. This method has been widely used in VAEs and GANs because it only requires a differentiable map from latent space to image space.
% \modify{However, it has limitations such as a lack of evidence for applying the Euclidean metric in image space. 이부분 저희 이번 논문에서 해결을 안해준것 같은데..ㅎ}
However, no studies have investigated the geometry of latent space of DMs utilizing the pullback metric.

%% file: 3method.tex
\section{Discovering the latent basis of DMs}
\label{sec:sec3}
In this section, we explain how to extract a latent structure of \exspace{} using differential geometry. 
First, we introduce a key concept in our method: the {\it pullback metric}. 
Next, by adopting the local Euclidean metric of \ehspace{} and utilizing the pullback metric, we discover the local latent basis of the \exspace{}. 
Moreover, although the direction we found is {\it local}, we show how it can be applied to other samples via parallel transport. 
Finally, we introduce $\vx$-space guidance for editing data in the \exspace{} {to enhance the quality of image generation.}

\subsection{Pullback metric} \label{sec:pullback}
% In this subsection, we explain how to extract a local latent basis of \exspace{}, and their corresponding local tangent basis of \ehspace{} using differential geometry. 
% First, we introduce a key concept in our method: the pullback metric. 
% The pullback metric is a mathematical tool to define a metric on a space by exploiting the metric structure of another metric space.
% Next, we adopt the local Euclidean metric of \ehspace{} to identify local semantic latent subspace for individual samples in \exspace{}. 

We consider a curved manifold, \exspace{}, where our latent variables $\mathbf{x}_t$ exist. 
The differential geometry represents $\mathcal{X}$ through patches of tangent spaces, $\tanxspace{}$, which are vector spaces defined at each point $\mathbf{x}$. 
Then, all the geometrical properties of $\mathcal{X}$ can be obtained from the inner product of $||d\mathbf{x}||^2 = \langle d{\mathbf{x}},d{\mathbf{x}} \rangle_\mathbf{x}$ in $\tanxspace{}$.
However, we do not have any knowledge of $\langle d{\mathbf{x}},d{\mathbf{x}} \rangle_\mathbf{x}$.
It is definitely not a Euclidean metric. Furthermore, samples of $\mathbf{x}_t$ at intermediate timesteps of DMs include inevitable noise, which prevents finding semantic directions in $\tanxspace{}$.

Fortunately, \citet{kwon2022diffusion} observed that \ehspace{}, defined by the bottleneck layer of the U-Net, exhibits local linear structure.
% Fortunately, \citet{kwon2022diffusion} observed that \ehspace{}, defined by the bottleneck layer of the U-Net, exhibits \yh{semantically} local linearity.
% Henceforth, we denote \ehspace{} as $\mathcal{H}$.
This allows us to adopt the Euclidean metric on $\mathcal{H}$.
In differential geometry, when a metric is not available on a space, {\it pullback metric} is used.
If a smooth map exists between the original metric-unavailable domain and a metric-available codomain, the pullback metric is used to measure the distances in the domain space.
Our idea is to use the pullback Euclidean metric on $\mathcal{H}$ to define the distances between the samples in $\mathcal{X}$.

DMs are trained to infer the noise $\mathbf{\epsilon}_t$ from a latent variable $\mathbf{x}_t$ at each diffusion timestep $t$. 
Each $\mathbf{x}_t$ has a different internal representation $\mathbf{h}_t$, the bottleneck representation of the U-Net, at different $t$'s.
The differentiable map between $\mathcal{X}$ and $\mathcal{H}$ is denoted as $f : \mathcal{X} \rightarrow \mathcal{H}$.
Hereafter, we refer to $\mathbf{x}_t$ as $\mathbf{x}$ for brevity unless it causes confusion. It is important to note that our method can be applied at any timestep in the denoising process.
%$f$ gives a linear map $\nabla_{\mathbf{x}} \mathbf{h} : \tanxspace{} \rightarrow \tanhspace{}$ between the domain and codomain tangent spaces.
{We consider a linear map, $\tanxspace{} \rightarrow \tanhspace{}$, between the domain and codomain tangent spaces.}
%The differential geometry then defines a linear map between the tangent space $\tanxspace{}$ at $\mathbf{x}$ and corresponding tangent space $\tanhspace{}$ at $\mathbf{h}$. The 
This linear map can be described by the {\it Jacobian} $J_{\mathbf{x}} = \nabla_{\mathbf{x}} \mathbf{h}$
which determines how a vector $\mathbf{v} \in \tanxspace{}$ is mapped into a vector $\mathbf{u} \in \tanhspace{}$ by
$\mathbf{u} = J_{\mathbf{x}} \mathbf{v}$. 
% In practice, the Jacobian can be computed from automatic differentiation of the U-Net.
% However, since the Jacobian of too many parameters is not tractable, we use a sum-pooled feature map of the bottleneck representation as our $\mathcal{H}$. 

Using the local linearity of $\mathcal{H}$, we assume the metric, $||d\mathbf{h}||^2 = \langle d\mathbf{h}, d\mathbf{h} \rangle_\mathbf{h} = d\mathbf{h}^{\tran} d\mathbf{h}$ as a usual dot product defined in the Euclidean space.
To assign a geometric structure to $\mathcal{X}$, we use the pullback metric of the corresponding $\mathcal{H}$.
In other words, the norm of $\mathbf{v} \in \tanxspace{}$ is measured by the norm of corresponding codomain tangent vector:
\begin{equation} \label{eq:pullback}
\begin{aligned}
||\dx{}||^2_\text{pb} \triangleq  \langle \dh{}, \dh{} \rangle_{\mathbf{h}} = \dx{}^{\tran} \jacx{}^{\tran} \jacx{}\dx{}.
\end{aligned}
\end{equation}

% \subsection{\yh{Find the local semantic latent directions}}
% \subsection{\yh{Extracting the semantic directions and editing}}
% \subsection{Extraction of semant directions}

% \yh{This subsection describes how we extract semantic latent directions using the pullback metric, and how we construct semantic latent subspace using discovered directions. The overall process is illustrated in \fref{fig:method_figure}.}

%%% ICML ver
% \label{sec:method_local}
% This subsection describes how we extract semantic latent directions using the pullback metric, and how we edit samples for multiple times given the meaningful directions by geodesic shooting. The overall process is illustrated in \fref{fig:method_figure}.

\begin{figure}[!t]
    \centering
    \includegraphics[width=1\linewidth]{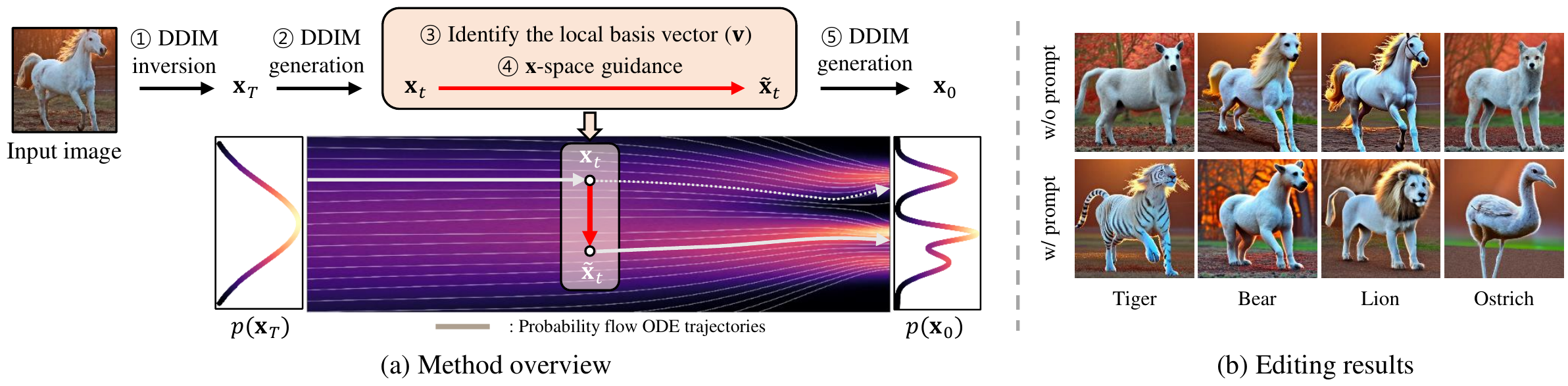}
    \vspace{-1em}
    \caption{\textbf{Image editing with the discovered latent basis.} 
    (a) Schematic depiction of our image editing procedure. 
    \modify{①} An input image is subjected to DDIM inversion, resulting in an initial noisy sample $\mathbf{x}_T$. 
    \modify{②} The sample $\mathbf{x}_T$ is progressively denoised until reaching the point $t$ through DDIM generation. 
    \modify{③} Subsequently, the local latent basis $\{\mathbf{v}_1, \cdots, \mathbf{v}_n\}$ is identified by using the pullback metric. 
    \modify{④} This enables the manipulation of the sample $\mathbf{x}_t$ along one of the basis vectors using $\mathbf{x}$-space guidance. 
    \modify{⑤} The DDIM generation concludes with the progression from the modified latent variable $\tilde{\mathbf{x}}_t$.
    (b) Examples of edited images using a selected basis vector. The latent basis vector \modify{could be} conditioned on prompts \modify{and it} facilitates text-aligned manipulations.
    }
    \vspace{-1em}
    \label{fig:method}
\end{figure}

\subsection{Finding local latent basis}
Using the pullback metric, 
{
we define the local latent vector $\mathbf{v} \in \tanxspace{}$ {that shows a large variability in $\tanhspace{}$}.
We find a unit vector $\mathbf{v}_1$ that maximizes $||\dx{}||^2_\text{pb}$.
}
% It can be interpreted as the first eigenvector of $\jacx{}^{\tran} \jacx{} = V \Lambda^2 V^{\top}$.
By maximizing $||\dx{}||^2_\text{pb}$ while remaining orthogonal to $\mathbf{v}_1$, one can obtain the second unit vector $\mathbf{v}_2$. This process can be repeated to have $n$ latent directions of $\{\mathbf{v}_1, \mathbf{v}_2, \cdots, \mathbf{v}_n \}$ in $\mathcal{T}_{\mathbf{x}}$. 
In practice, $\mathbf{v}_i$ corresponds to the $i$-th right singular vector from the singular value decomposition {(SVD)} of $\jacx{} = U \Lambda V^{\tran}$, i.e., $J_{\vx} \vv_{i} = \Lambda_{i} \vu_{i}$. 
Since the Jacobian of too many parameters is not tractable, we use a \textit{power method} \cite{golub2013matrix,miyato2018spectral,haas2023discovering} to approximate the SVD of $\jacx{}$
{(See \aref{appendix:comparisons} for the time complexity and \aref{appendixsec:algorithm} for the detailed algorithm).} 

{Henceforth, we refer to $\mathcal{T}_{\mathbf{x}}$ as a local latent subspace, and $\{\mathbf{v}_1, \mathbf{v}_2, \cdots, \mathbf{v}_n \}$ as the corresponding local latent basis.}
\begin{align}
    \mathcal{T}_{\mathbf{x}} \triangleq \text{span} \{\mathbf{v}_1, \mathbf{v}_2\, \cdots, \mathbf{v}_n\},~\text{where}~\mathbf{v}_i~\text{is}~i\text{-th right singular vector of}~J_{\mathbf{x}}.
\end{align}

%%% ICML ver : construct Th in h space 
Using the linear transformation between $\tanxspace{}$ and $\tanhspace{}$ via the Jacobian $\jacx{}$, one can also obtain corresponding directions in $\mathcal{T}_{\mathbf{h}}$.
% \begin{align}
%     \vu{}_i = \frac{1}{\lambda_i} \jacx{} \mathbf{v}_i.
% \end{align}
% Here, we normalize $\vu{}_i$ by dividing the $i$-th singular value $\lambda_i$ of \jo{the diagonal matrix} $\Lambda$ to preserve the Euclidean norm $||\mathbf{u}_i||=1$. 
In practice, $\mathbf{u}_i$ corresponds to the $i$-th left singular vector from the {SVD} of $\jacx{}$.
After selecting the top $n$ (e.g., $n = 50$) directions of large eigenvalues, we can approximate any vector in $\mathcal{T}_{\mathbf{h}}$ with {a} finite basis, $\{\mathbf{u}_1, \mathbf{u}_2, \cdots, \mathbf{u}_n \}$.
When we refer to a local tangent space henceforth, it means the $n$-dimensional low-rank approximation of the original tangent space.
\begin{align} \label{eq:local_tangent_space}
    \mathcal{T}_{\vh} \triangleq \text{span} \{\vu_1, \vu_2\, \cdots, \vu_n\},~\text{where}~\vu_i~\text{is the}~i\text{-th left singular vector of }~J_{\vx}.
\end{align}

%% mingi ver
% \paragraph{\yh{Interpretation of local latent basis}}
%The set of local latent basis vectors, $\{\vv_1, \vv_2, \cdots, \vv_n\}$, obtained through our proposed method, can be seen as a `{\it signal}' that the model is particularly sensitive given $\vx$.
%% These vectors represent the directions in which the model's representation $\mathbf{h}$ undergoes the most significant changes, revealing the information that the model focuses on from $\mathbf{x}_t$ for a specific sample.
%On the other hand, the basis of local tangent space, represented as $\{\vu_1, \vu_2\, \cdots, \vu_n\}$, can be thought of as the corresponding `{\it representation}' linked to the signal.
{The collection of local latent basis vectors, $\{\vv_1, \vv_2, \cdots, \vv_n\}$, obtained through our proposed method, can be interpreted as a {\it signal} that the model is highly response to for a given $\vx$.
On the other hand, the basis of the local tangent space, denoted as $\{\vu_1, \vu_2\, \cdots, \vu_n\}$, can be viewed as the corresponding {\it representation} associated with the signal.
}

{In Stable Diffusion, the prompt also influences the Jacobian, which means that the local basis also depends on it. }
% We can utilize any prompt to obtain a local latent basis.
{
We can utilize any prompt to obtain a local latent basis, and different prompts create distinct geometrical structures.
}
% \mingi{In a conditional model, the local tangent space undergoes prompt-dependent modifications, allowing us to derive a specific basis by using the prompt as a condition of the model. Note that we can utilize any prompt to obtain a local latent basis.}
{For the sake of brevity, we will omit the word {\it local} unless it leads to confusion.}

\subsection{{Generating edited images with $\mathbf{x}$-space guidance}}
% \modify{찾은 basis를 어떻게 사용하는 건지에 대한 이야기임을 말하면서 글 다듬기.}
{A na\"ive approach for manipulating a latent variable $\mathbf{x}$ using a latent vector $\mathbf{v}$ is through simple addition, specifically $\mathbf{x}+\gamma\mathbf{v}$.
However, using the na\"ive approach sometime leads to noisy image generation.
% because the basis vectors are found in the space \exspace{} where the latent variables $\mathbf{x}$, that are prone to noise, exist.
To address this issue, instead of directly using the basis for manipulation, we use a basis vector that has passed through the decoder once for manipulation.
}
%To manipulate the latent variable $\mathbf{x}$ with a given local latent basis vector $\mathbf{v}$, the simple approach is adding basis to the latent variable, i.e. $\mathbf{x}+\gamma\mathbf{v}$. 
%However, \exspace{}, where we find the basis, is a noisy space by default and simple approach sometimes gives rise to noisy results. 
%To solve this problem, we use a basis vector for manipulation after going through the decoder once, instead of using the basis directly for manipulation.
The $\mathbf{x}$-space guidance is defined as follows
\begin{align}
\label{eq:xguidance}
    \tilde{\mathbf{x}}_{\text{XG}} = \mathbf{x} + \gamma[\epsilon_{\theta}(\mathbf{x}+\mathbf{v}) - \epsilon_{\theta}(\mathbf{x})]
\end{align}
where $\gamma$ is a hyper-parameter controlling the strength of editing and $\epsilon_{\theta}$ is a diffusion model. 
% \jo{(((g는 함수인가?)))}
Equation~\ref{eq:xguidance} is inspired by classifier-free guidance, but the key difference is that it is directly applied in the latent space \exspace{}.
{Our $\mathbf{x}$-space guidance provides qualitatively similar results to direct addition, 
\modify{while} it shows better fidelity. (See \aref{appendixsec:ablation_x_guidance} for ablation study.)}
% \yh{We provide an ablation study on this in \sref{}.}

\subsection{The overall process of image editing}
In this section, we summarize the entire editing process with five steps: 1) The input image is inverted into initial noise $\mathbf{x}_T$ using DDIM inversion. 2) $\mathbf{x}_T$ is gradually denoised until $t$ through DDIM generation. 3) Identify local latent basis $\{ \mathbf{v}_1, \cdots, \mathbf{v}_n \}$ using the pullback metric at $t$. 4) Manipulate $\mathbf{x}_t$ along the one of basis vectors using the $\mathbf{x}$-space guidance. 5) The DDIM generation is then completed with the modified latent variable $\tilde{\mathbf{x}}_t$. \fref{fig:method} illustrates the entire editing process. 

In the context of a text-to-image model, such as Stable Diffusion, it becomes possible to include textual conditions while deriving local basis vectors. 
% 이를 통해 DDIM inversion 이나 generation 때 text 에 의한 guidance 를 주지 않고도 text condition 에 맞는 editing 을 할 수 있게 된다.
\modify{
Although we do not use any text guidance during DDIM inversion and generation, a local basis with text conditions enables semantic editing that matches the given text.
% It enables semantic editing that matches the text conditions even without giving guidance based on text during DDIM inversion or generation.
}
% It aligns all the local basis vectors with the condition text. 
Comprehensive experiments can be found in Section 4.1.

% It is noteworthy that our approach \modify{needs no extra training}. Moreover, it achieves semantically meaningful image editing, only involving a traversal of the latent variable within a single timestep.
\modify{
It is noteworthy that our approach needs no extra training and simplifies image editing by only adjusting the latent variable within a single timestep.
}

\subsection{{Editing various samples with parallel transport}}

\modify{
Let us consider a scenario where our aim is to edit ten images, changing straight hair to curly hair. 
Due to the nature of the unsupervised image editing method, it is becomes imperative to manually check the semantic relevance of the latent basis vector in the edited results. 
% Considering the inherent characteristics of the unsupervised image editing method, it becomes imperative to manually inspect the semantic relevance of the latent basis vector within the edited results. 
Thus, to edit every samples, we have to manually find a straight-to-curly basis vector for individual samples.
% Therefore, for each editing task, we must manually identify a straight-to-curly basis vector specific to each image.
}

\modify{
One way to alleviate this tedious task is to apply the curly hair vector obtained from one image to other images. 
}
However, the basis vector $\vv \in \tanxspace{}$ obtained at $\vx$ cannot be used for the other sample $\vx'$ because $\vv \notin \mathcal{T}_{\vx'}$. 
% 왜냐면, 
% 이 번거로운 작업을 거쳐야 하는 이유는 우리가 local basis 를 찾았기 때문이다. 
% 우리가 a latent basis of $\tanxspace{}$ 에서 curly hair 방향을 찾더라도 
% So far, we extracted a latent basis of $\tanxspace{}$ and a tangent basis of $\tanhspace{}$ given {a} latent variable $\vx$. 
% Thus, 우리가 구한 direction 을 다른 sample 에 적용하기 위해선 it is necessary to relocate the extracted direction to a new tangent space.
% However, the basis vector obtained cannot be used for the other sample $\vx'$ because $\vv \notin \mathcal{T}_{\vx'}$. 
% If we do not use P.T., we have to manually find a straight-to-curly basis vector for individual samples. P.T. allows transporting a straight-to-curly basis vector in one sample to all other samples to edit all images with only one manual inspection.
Thus, in order to apply the direction we obtained to another sample, it is necessary to relocate the extracted direction to a new tangent space.
To achieve this, we use parallel transport that moves $\vv_i$ onto the new tangent space $\mathcal{T}_{\vx'}$.

Parallel transport moves {a tangent vector $\vu \in \tanhspace{}$ to $\vu' \in \mathcal{T}_{\vh'}$} without changing its direction as much as possible while keeping the vector tangent on the manifold \cite{shao2018riemannian}.
{It is notable that the parallel transport in curved manifold significantly modifies the original vector.
% It is notable that the projection significantly modifies the original vector $\mathbf{v}_i \in \tanxspace{}$, because $\mathcal{X}$ is a curved manifold.
Fortunately,} $\mathcal{H}$ is relatively flat. Therefore, it is beneficial to apply the parallel transport in $\mathcal{H}$.

We aims to move $\vv_i$ onto new tangent space $\mathcal{T}_{\mathbf{x}'}$, using parallel transport in $\mathcal{H}$.
First, we convert the latent direction $\vv_i \in \mathcal{T}_{\mathbf{x}}$ to the corresponding direction of $\vu_i \in \mathcal{T}_{\mathbf{h}}$.
Second, we apply the parallel transport $\vu_i \in \mathcal{T}_{\mathbf{h}}$ to ${\mathbf{u}'}_{i} \in \mathcal{T}_{\mathbf{h}'}$, where $\mathbf{h}' = f(\mathbf{x}')$. 
% Specifically, we consider the parallel transport along a geodesic between $\vh$ to $\vh'$.
In the general case, parallel transport involves iterative projection and normalization on the tangent space along the path connecting two points \cite{shao2018riemannian}.
However, in our case, we assume that \ehspace{} has Euclidean geometry. 
Therefore, we move $\vu$ directly onto $\mathcal{T}_{\mathbf{h}'}$ through projection, without the need for an iterative process.
Finally, transform $\mathbf{u}'_{i}$ into $\mathbf{v}'_i \in \mathcal{X}$.
% % Third, we obtain $\mathbf{v}'_{i}$ by transforming $\mathbf{u}'_{i}$ into $\mathcal{X}$.
Using this parallel transport of $\mathbf{v}_i \to \mathbf{v}'_i$ via $\mathcal{H}$, 
we can apply the local latent basis obtained from $\vx$ to edit or modify the input $\vx'$.

%% file: 4experiment.tex
\begin{figure}[!t]
    \centering
    % \vspace{-2em}
    \includegraphics[width=1.0\linewidth]{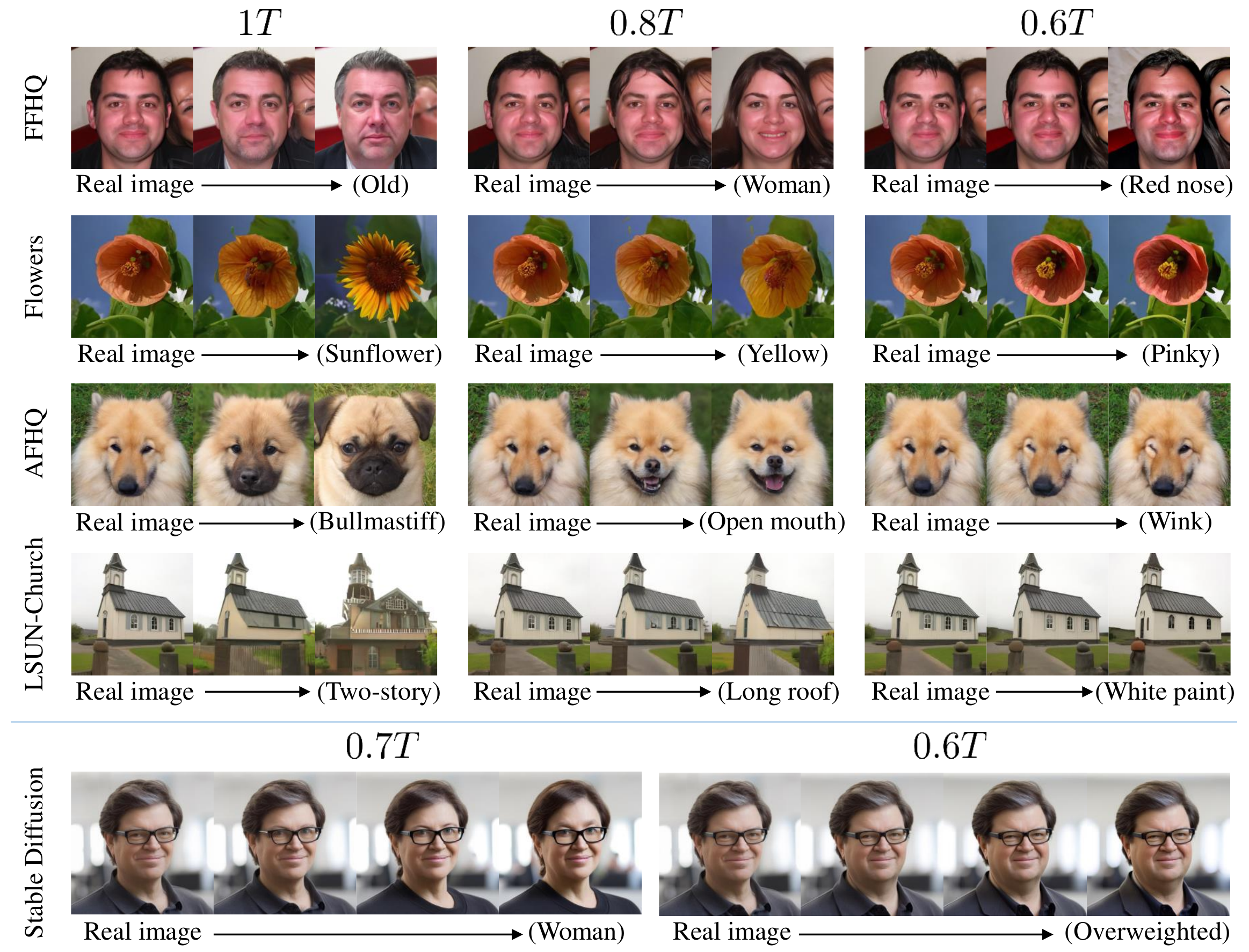}
    \vspace{-1em}
    \caption{\textbf{{Examples of image edition using the latent basis}.} 
    The attributes are manually interpreted {since the editing directions are not intentionally supervised.}
    For Stable Diffusion, we {used} an empty string as a prompt.
    {Each column represents edits made} at different diffusion timesteps ($0.6T$, $0.8T$, and $T$ for the unconditional diffusion model; $0.6T$ and $0.7T$ for Stable Diffusion).
    }
    \vspace{-1em}
    \label{fig:local_basis}
\end{figure}

\begin{figure}[!t]
    \centering
    % \vspace{-1.0em}
    \includegraphics[width=1\linewidth]{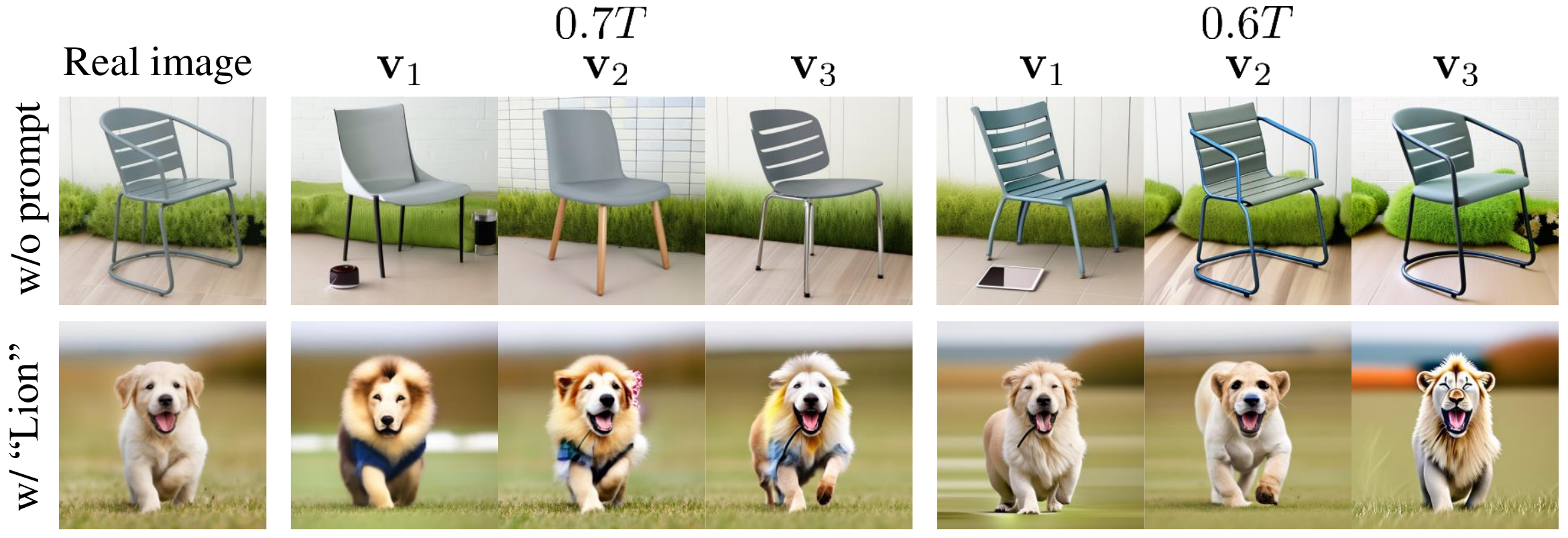}
    \vspace{-1.5em}
    \caption{
    \textbf{{Examples of image edition using top-3 latent basis vectors.}} 
    {Each column is edited using a} different latent vector 
    $\{ \vv_1, \vv_2, \vv_3\}$. % are the results of manipulating the top three latent basis vectors, respectively. 
    Each group of columns {represents edits made} at different diffusion timesteps ($0.6T$ and $0.7T$).
    {Notably, when} given the ``Lion" prompt, {it is evident that all the top latent basis vectors align with the direction of the prompt.}
    %we can see that the top latent basis vectors all point in the direction of the prompt.
    }
    \vspace{-0.5em}
    \label{fig:local_basis_text_vk}
\end{figure}

\begin{figure}[!t]
    \centering
    \includegraphics[width=1\linewidth]{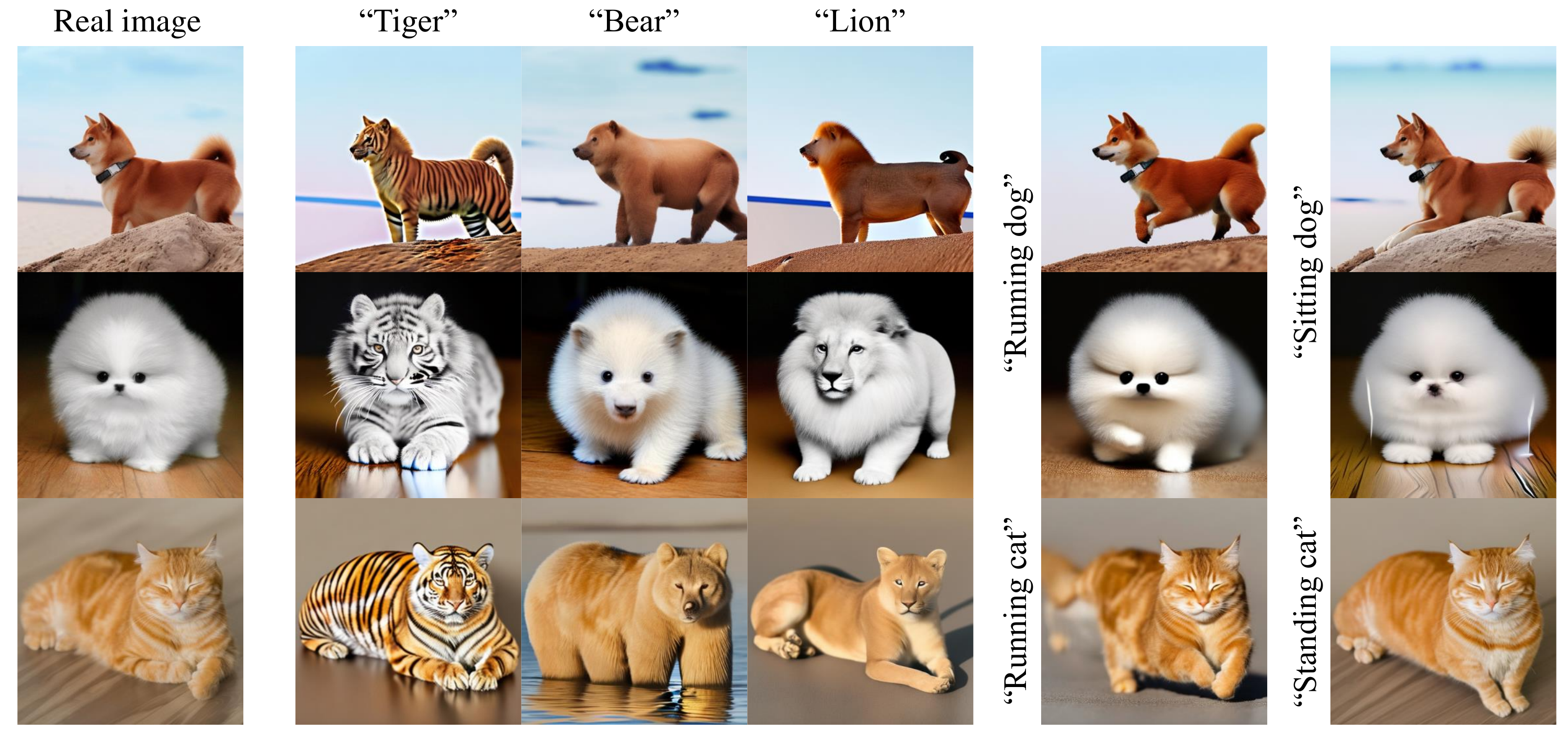}
    \vspace{-1.5em}
    \caption{\textbf{Examples of image edition using latent basis vectors discovered with various prompts.}
    {Each column is edited using the latent basis vector obtained from a different text prompt. Importantly, our method employs each prompt only once to derive the local latent basis.}
    %Different columns are edited by the latent basis vector found given a different text. It is worth noting that our method only uses a prompt once to get the local latent basis.
    % when manipulates the latent variable $\vx_t$.
    % \modify{샘플링할때 텍스트 영향은 아예 없는거고, 이에대한 설명 좀만 추가.}
    }
    \vspace{-1em}
    \label{fig:local_basis_text}
\end{figure}

\section{Findings and results}
In this section, we analyze the geometric structure of DMs with our method.
% In \sref{sec:local}, we demonstrate the potential of the local latent basis for real image editing.
\sref{sec:local} demonstrates that the latent basis found by our method can be used for image editing.
% In \sref{sec:local}, we validate our method by demonstrating the potential of the latent basis for real image editing.
In \sref{sec:evolution}, we investigate how the geometric structure of DMs evolves as the generative process progresses. 
% \textcircled{\raisebox{-0.9pt}{1}} The frequency domain of local latent basis changes from coarse to fine feature as the generative process progresses. % (\sref{sec:evolution-x})
% \textcircled{\raisebox{-0.9pt}{2}} The differences of local tangent spaces from various samples getting larger as the generative process progresses. % and show the possibility of parallel transport on a local basis. (\sref{sec:evolution-samples})
% \textcircled{\raisebox{-0.9pt}{3}} The differences of local tangent spaces from various timesteps depends on the complexity of the dataset.
%Lastly, in \sref{sec:text}, we discover that the geometric structure the geometric properties of the text-condition model change with a given text. % is provided.
Lastly, in \sref{sec:text}, we examine how the geometric properties of the text-condition model change with a given text. % is provided.

{The implementation details of our work are provided on \aref{appendix:implementation_detail}}. The source code for our experiments is included in the supplementary materials and will be publicly available upon publication.

\subsection{Image editing with the latent basis}
% \subsection{\jo{latent-based editing with and without prompts}}
\label{sec:local}
% \yh{TODO : local figure 완성되고 나서 작성}

% In \sref{sec:sec3}, we introduce a method for determining local latent basis utilizing the pullback metric. 
In this subsection, we demonstrate the capability of our discovered latent basis for image editing. 
To extract the latent variables from real images for editing purposes, we use DDIM inversion.
In experiments with Stable Diffusion (SD), we do not use guidance, i.e., unconditional sampling, for both DDIM inversion and DDIM sampling processes.
This ensures that our editing results solely depend on the latent variable, not on other factors such as prompt conditions.
% This ensures that our editing result is solely the outcome of manipulating the latent variable.
% Consequently, we only utilize the text guide to extract the local latent structure and keep the generative process as unconditional sampling to minimize the influence of text.
% It is worth to note that we only use the text condition when we extract the local latent structure
% when given a text (refer to \fref{fig:text}). 

% \modify{add text conditional basis; cross attention이 특정 feature를 강조할텐데 그걸 pc를 구해도 다 텍스트에 관련된 basis만 나오는 것을 기대해야 하고, 우리가 관측한것도 이런 기대에 부합한다. text cond 피규어는 여러 local basis중에 하나 고른거고 나머지는 appendix봐라.}

Figures~\ref{fig:method} and \ref{fig:local_basis} illustrate the example results edited by the latent basis found by our method.
% demonstrate the discovery of diverse basis across different datasets which highlights the superiority of our approach \emph{without supervision} such as CLIP or a classifier. 
The latent basis clearly contains semantics such as age, gender, species, structure, and texture. 
% When modulating the strength of the basis, the resulting image undergoes continuous changes, as exemplified in \fref{fig:local_basis}. 
{Note that editing at timestep $T$ yields coarse changes such as age and species. On the other hand, editing at timestep $0.6T$ leads to fine changes, such as nose color and facial expression.} 
% Appendix provides more examples.
% In the subsequent section, we delve into a detailed analysis of the latent basis from these distinct timesteps.

% Significantly, editing at timestep $T$ results in coarse changes, whereas editing at timestep $0.6T$ produces finer changes, as exemplified by various examples. 
% In the following section, we will thoroughly explore and analyze the local semantic foundation derived from these distinct timesteps.

\fref{fig:local_basis_text_vk} demonstrates the example results edited by the various latent basis vectors.
Interestingly, using the text ``lion'' as a condition, 
the entire latent basis {captures lion-related attributes.} % aligns with the text.
% Moreover, \fref{fig:local_basis_text} shows that the latent basis can be aligned with the text not only for the type of object but also the action.
Furthermore, \fref{fig:local_basis_text} shows that the latent basis aligns with the text not only in terms of object types but also in relation to pose or action.
\modify{For a qualitative comparison with other state-of-the-art image editing methods, refer to 
\aref{appendix:comparisons}. For more examples of editing results, refer to \aref{appendix:additional_results}.}

\subsection{{Evolution of latent structures during {generative} processes}}
\label{sec:evolution}

\begin{wrapfigure}{r}{6cm}
    \vspace{-1em}
\includegraphics[width=6cm]{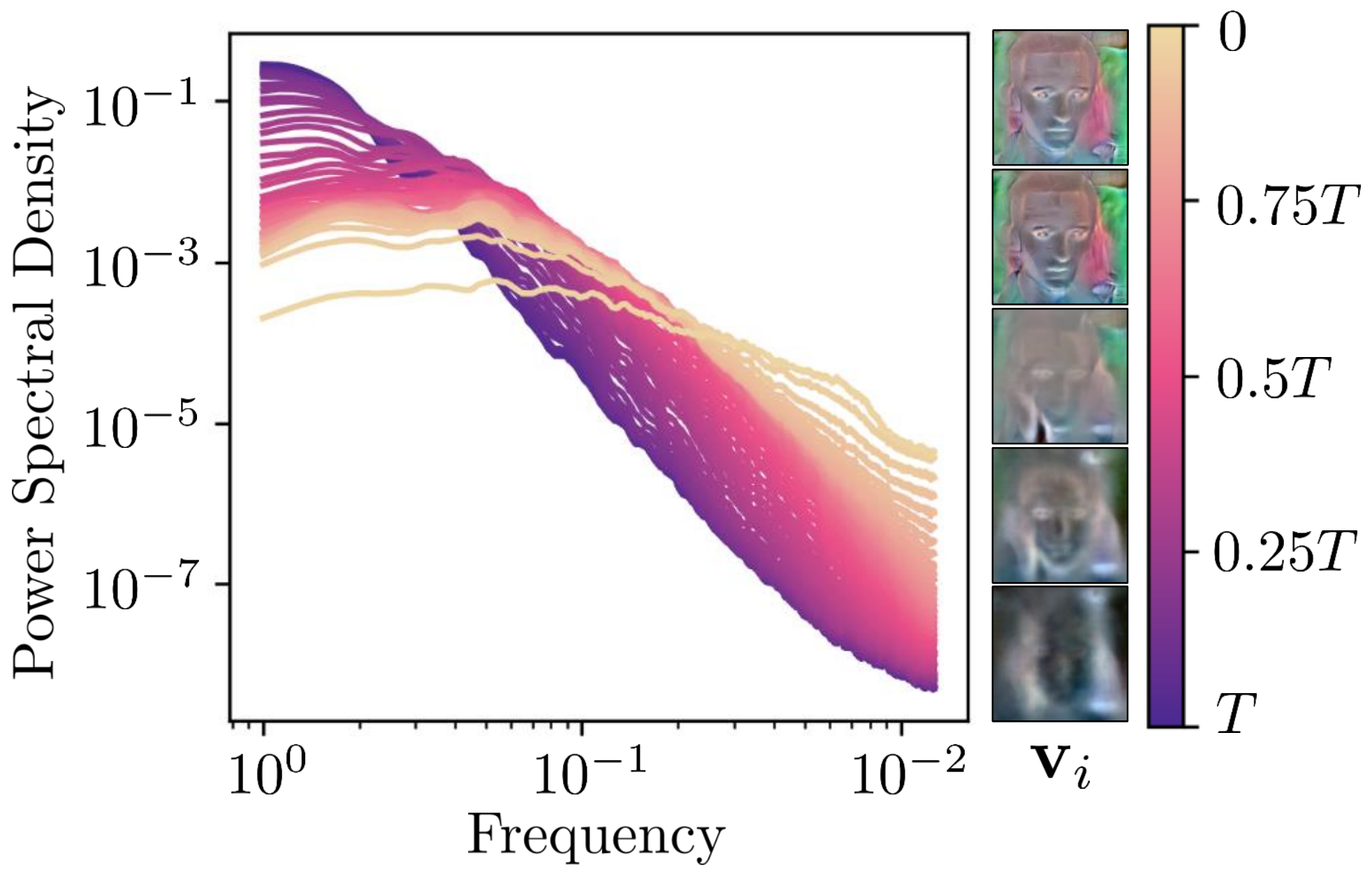}
    \vspace{-1.5em}
    \caption{\textbf{{Power Spectral Density (PSD) of latent basis.}} {The PSD at $t = T$ (purple) exhibits a greater proportion of low-frequency signals, while the PSD at smaller $t$ (beige) reveals a larger proportion of high-frequency signals. The latent vectors $\vv_i$ are min-max normalized for visual clarity.}
    %\textbf{Power Spectral Density of latent basis}
    %The PSD at $t = T$ (purple line) shows a larger portion of
    %low-frequency signals, whereas the PSD at smaller $t$ (beige line) shows a larger portion of high-frequency signals. The latent vector $\vv_i$ are min-max normalized for visual purposes.
    }\label{figure:PSD}
    \vspace{-2em}
\end{wrapfigure}

In this subsection, we demonstrate how the latent structure evolves during the generative process and identify three trends. 
%\textcircled{\raisebox{-0.9pt}{1}} 
1) {The frequency domain of the latent basis changes from low to high frequency. It agrees with the previous observation that DMs generate samples in coarse-to-fine manner.} % The frequency domain of latent basis changes from coarse to fine feature as the generative process progresses. % (\sref{sec:evolution-x})
%\textcircled{\raisebox{-0.9pt}{2}} 
2) The difference between the tangent spaces of different samples increases over the generative process. It implies finding generally applicable editing direction in latent space becomes harder in later timesteps.
 % and show the possibility of parallel transport on a local basis. (\sref{sec:evolution-samples})
%\textcircled{\raisebox{-0.9pt}{3}} 
3) The differences of tangent spaces {between} timesteps depend on the complexity of the dataset.

% We examine the frequency domain of the local latent basis from different timesteps using the power spectral density (PSD), and find that DMs become more attuned to finer features as the process progresses. 
% We also investigate variations in the local tangent space using the geodesic metric, showing that simpler data leads to greater similarity in the local tangent subspace at different timesteps. This suggests that simpler data leads to increasingly similar features processed at each timestep.
% 이 관찰으로부터 서로 다른 timestep 유사한 signal 을 처리하지 않도록 timestep 을 배치하니 uniform timestep 에 비해 더 좋은 sampling fidelity 를 얻을 수 있었다. 

\paragraph{{Latent bases gradually evolve from low- to high-frequency structures.}}
%\paragraph{latent basis evolves from low frequencies to high frequencies.}
% \paragraph{The frequency domain of latent basis evolves from low-to-high.}
\label{sec:evolution-x}

% In \sref{sec:local}, we observed that coarse manipulations are performed at the beginning of the denoising step, and then finer manipulations are performed as the timestep progresses. 
% To quantitatively verify this observation, we plotted the power spectral density (PSD) of the discovered latent basis. 

\fref{figure:PSD} is the power spectral density (PSD) of the discovered {latent} basis over various timesteps. 
The early timesteps contain a larger portion of low frequency than the later timesteps and the later timesteps contain a larger portion of high frequency.

This suggests that the model focuses on low-frequency signals at the beginning of the generative process and then shifts its {focus} to high-frequency signals over time.
This result strengthens the common understanding
about the coarse-to-fine behavior of DMs over the generative process \cite{choi2022perception, daras2022multiresolution}.

% \yh{This result strengthens the common understanding that DMs start to create coarse features and gradually refine them over a generative process.}
% This suggests that the model focuses on low-frequency signals at the beginning of the generative process and then shifts its attention to high-frequency signals over time.
% long version
% \yh{
% In the forward diffusion process, high-frequency details are perturbed faster than low-frequency signals. Therefore, as the diffusion timestep approaches $t \approx T$, only the low-frequency signals remain in the image. Hence, it is reasonable for a model to reconstruct the original image by focusing on the remaining low-frequency signals.
% }
% short version
% \yh{
% Considering high-frequency perturbed faster in forward diffusion process, it is natural for model to focus on low-frequency signal when $t \approx T$.
% }
% The PSD plots for additional datasets are provided in \aref{appendix:PSD}. Note that each dataset demonstrates a consistent qualitative trend.

% \begin{wrapfigure}{R}{5cm}
% \vspace{-2em}
% \includegraphics[width=5cm]{figure/Homogenity_samples.pdf}
% \vspace{-2em}
% \caption{
% \textbf{Geodesic distance across tangent space of different samples in various diffusion timesteps}
% Each point represents the average geodesic distance between pairs of 15 samples. 
% The tangent spaces of different samples are only similar early in the generative process.
% }
% \label{homogenity-sample}
% % \vspace{-2em}
% \label{fig:homo}
% \end{wrapfigure}

\paragraph{{The discrepancy of tangent spaces from different samples increases along the generative process.}}
%\paragraph{The difference between the tangent spaces from various samples increases as the generative process progress.}
\label{sec:evolution-samples}

% In our previous findings, we investigated the evolution of the latent basis. The next step is to focus on the evolution of the basis's corresponding {\it ``representation''}, i.e. tangent space. % original 아래
% From now on, we are going to turn our attention to the evolution of the  corresponding {\it ``representation''}, i.e. tangent space.

% In our previous findings, we established that the signal frequency in DM undergoes changes depending on the timestep. In order to gain further insights, we will now examine the representation of the local basis. 

% It is important to emphasize that the local basis is extracted from each local tangent space. This allows us to utilize the geodesic metric to assess the similarity between the discovered local subspaces across various samples, considering the local tangent subspaces generated at different timesteps.
%As a starting point, we checked how similar the discovered local subspaces of the different samples are to each other. 
% To begin, we examined the similarity between the discovered local subspaces across different samples.
% As a starting point, we checked how similar the local tangent spaces $\mathcal{T}_{\mathbf{h}}$ of the different samples are to each other. 
% The distortion of different vector spaces can be measured by the Grassman metric. 

% \begin{wrapfigure}{r}{3.2cm}
% % \vspace{-3em}
% \includegraphics[width=3.2cm]{figure/geodesic_metric.pdf}
% \vspace{-1em}
% \caption{Conceptual illustration of geodesic metric}\label{fig:geodesic_metric}
% \vspace{-1.25em}
% \label{fig:homo}
% \end{wrapfigure}

To investigate the geometry of the tangent basis, we employ a metric on the Grassmannian manifold. The Grassmannian manifold is a manifold where each point is a vector space, and the metric defined above represents the distortion across various vector spaces.
% By employing a metric on the Grassmannian manifold, one can determine the extent of distortion across various vector spaces.
We use \emph{geodesic metric}~\cite{choi2021not, ye2016schubert} to define the discrepancy between two subspaces $\{\mathcal{T}^{(1)}, \mathcal{T}^{(2)}\}$: 

% The distortion of different vector spaces can be found by the Grassman metric. We use \emph{geodesic metric}~\cite{choi2021not, ye2016schubert} to define the distortion between two local latent subspace centered at $\{\vh^{(1)}, \vh^{(2)}\}$: 

% \begin{figure}[h]
%    \centering
%    \vspace{-10pt}
%    \includegraphics[height=40pt,right]{figure/geodesic_distance_mini.pdf}
%    % \vspace{-10pt}
%    \vspace{-60pt}%
%    \label{fig:cclogo}
% \end{figure}

%%% Small Geodesic metric figure
% \begin{figure}[h]
%    \centering
%    \vspace{-10pt}
%    \includegraphics[height=30pt,right]{figure/geodesic_metric.pdf}
%    % \vspace{-10pt}
%    \vspace{-60pt}%
%    \label{fig:cclogo}
% \end{figure}

\begin{equation}
D_{\text{geo}}(\mathcal{T}^{(1)}, \mathcal{T}^{(2)}) = \sqrt{\sum_k \theta_k^2},
\end{equation}

\begin{wrapfigure}{r}{6cm}
\includegraphics[width=5cm, center]{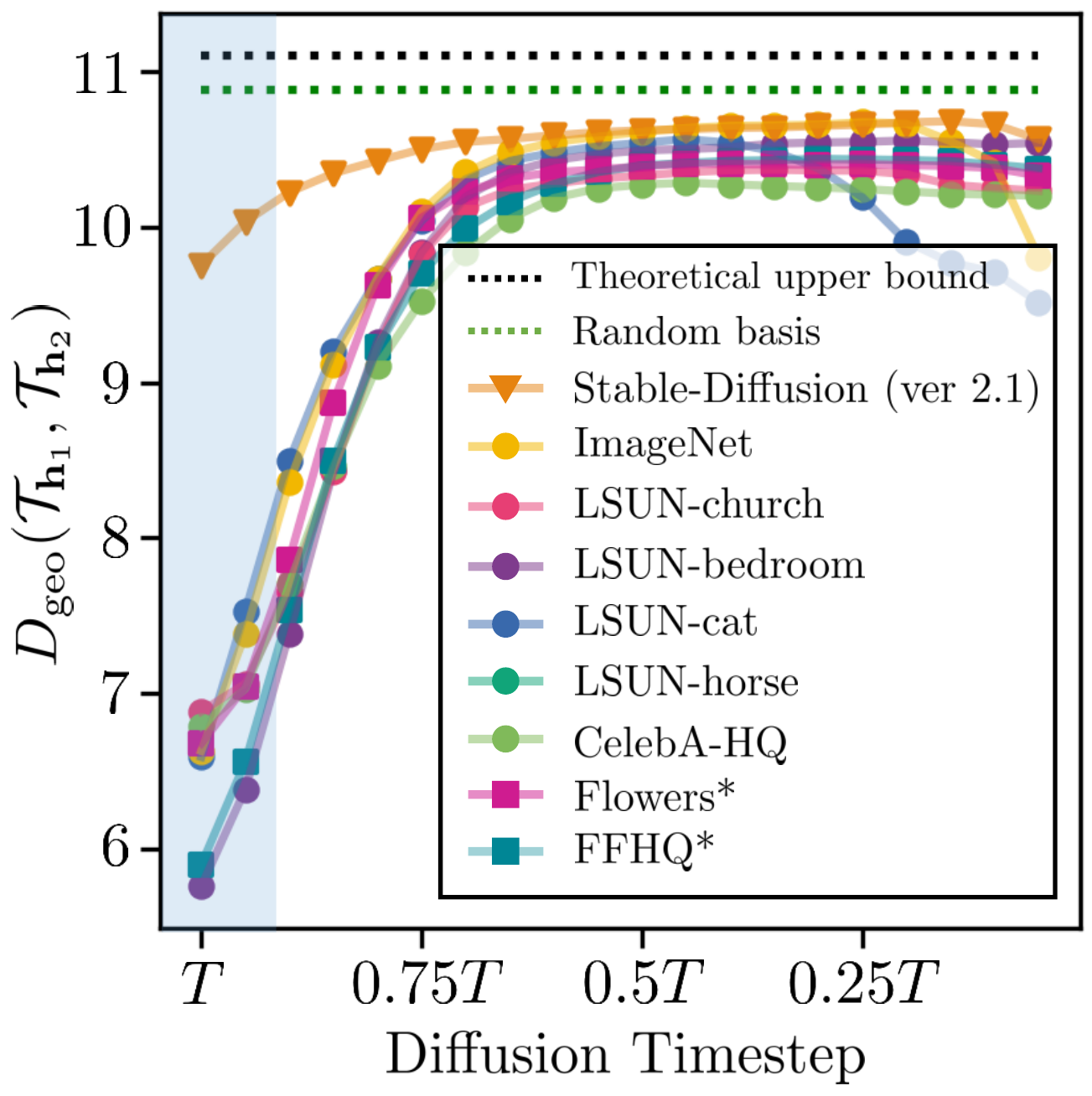}
    \vspace{-1em}
    \caption{
    \textbf{Geodesic distance across tangent space of different samples {at} various diffusion timesteps.}
    Each point represents the average geodesic distance between pairs of 15 samples.
    { It is notable that the similarity of tangent spaces among different samples diminishes as the generative process progresses.}
    %The tangent spaces of different samples are only similar early in the generative process.
    }
    \label{homogenity-sample}
\vspace{-3em}
\label{fig:homo}
\end{wrapfigure}

\begin{figure}[!b]
    \centering
    \includegraphics[width=1\linewidth]
    {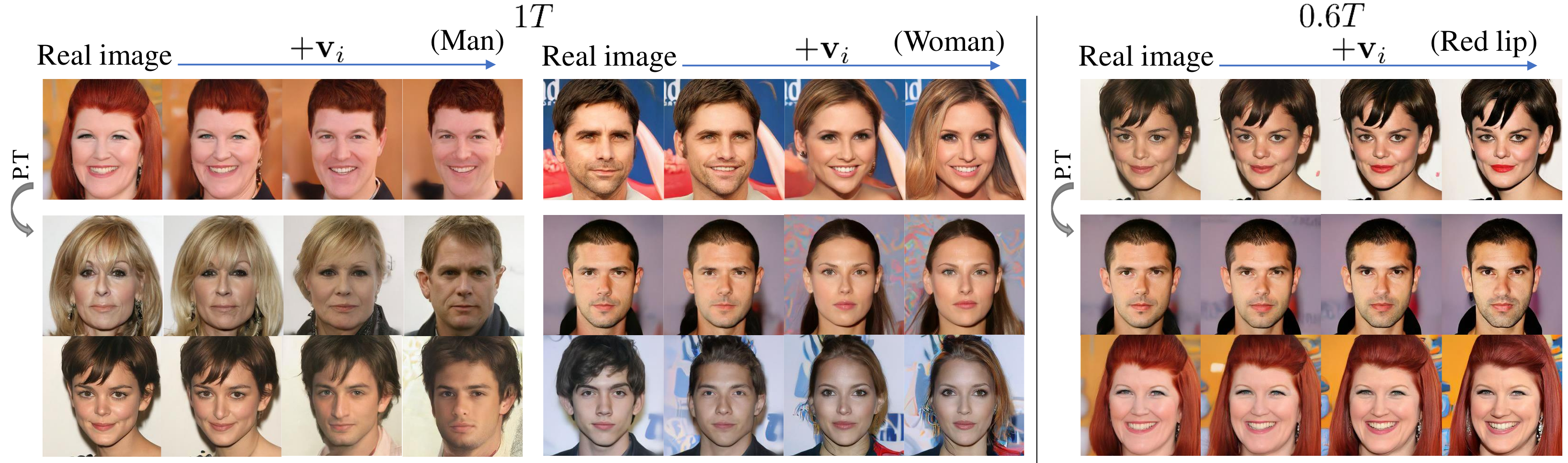}
    % \vspace{-1em}
    \caption{\textbf{\jo{Examples of image edition using parallel transport.}}
    \jo{The first row demonstrates the results of editing with their respective latent vectors, while the subsequent rows exhibit the results of editing through the parallel transport (P.T) of the latent vectors used in the first row. The latent vector performs effectively when $t = T$ (left and middle), but comparatively less satisfactorily for $0.6T$ (right).}
    %The first row show the result of editing with their own latent vector, and the rest of the rows show the result of editing with a parallel transport of the first row. The latent vector works well for when $t=T$ (left, middle), but relatively poorly for $0.6T$ (right). 
    }
    % \vspace{-1em}
    \label{fig:parallel}
\end{figure}

where {$\theta_k$} denotes the $k$-th principle angle between $\mathcal{T}^{(1)}$ and $\mathcal{T}^{(2)}$. Intuitively, the concept of geodesic metric can be understood as an angle between two vector spaces.
%(\ifref{fig:geodesic_metric})
{Here, the comparison between two different spaces was conducted for 
$\{\mathcal{T}_{\vh_1}, \mathcal{T}_{\vh_2} \}$. 
Unlike the \exspace{}, the \ehspace{} assumes a Euclidean space which makes the computation of geodesic metric that requires an inner product between tangent spaces easier. 
The relationship between tangent space and latent subspace is covered in more detail in \aref{appendixsec:relationship_tangent_latent}.
}

% \begin{wrapfigure}{r}{6cm}
%     \vspace{-1em}
% \includegraphics[width=6cm]{figure/PSD.pdf}
%     \vspace{-1.5em}
%     \caption{\textbf{{Power Spectral Density (PSD) of latent basis.}} {The PSD at $t = T$ (purple) exhibits a greater proportion of low-frequency signals, while the PSD at smaller $t$ (beige) reveals a larger proportion of high-frequency signals. The latent vectors $\vv_i$ are min-max normalized for visual clarity.}
%     %\textbf{Power Spectral Density of latent basis}
%     %The PSD at $t = T$ (purple line) shows a larger portion of
%     %low-frequency signals, whereas the PSD at smaller $t$ (beige line) shows a larger portion of high-frequency signals. The latent vector $\vv_i$ are min-max normalized for visual purposes.
%     }\label{figure:PSD}
%     \vspace{+2em}
% \end{wrapfigure}

\fref{fig:homo} demonstrates that the tangent space\uh{s} of the different samples are the most similar at $t=T$ and diverge as {timestep becomes} zero. 
% This aligns with the existing understanding that, after a certain timestep, the reverse diffusion trajectories of different samples are isolated from each other\cite{zhang2022gddim, wu2022uncovering}.

Moreover, the similarity across tangent space{s} allows us to effectively transfer the latent basis from one {sample} to another through parallel transport as shown in \fref{fig:parallel}.
% as showcased in \sref{sec:parallel}. 
% \fref{fig:parallel} demonstrates the edited results of parallel transporting the latent basis obtained from one sample to another sample.
% In $T$, where homogenity of tangent space exists, 
In $T$, {where the tangent spaces are homogeneous},
we consistently obtain semantically aligned editing results. 
{On the other hand, parallel transport at $t=0.6T$ does not lead to satisfactory editing because the tangent spaces are hardly homogeneous. Thus, we should examine the similarity of local subspaces to ensure consistent editing across samples.}
% However, when attempting parallel transport at $t=0.6T$, where homogeneity is lacking, it fails to produce satisfactory outcomes.
% This underscores the significance of considering the similarity of local subspaces.

% This is corroborated by the results depicted in \fref{fig:parallel}. 
% By transferring the local semantic direction from the first sample to subsequent samples, we consistently obtain semantically aligned editing results. However, it is important to note that when attempting to transfer the local basis at $t=0.6T$, where homogeneity is lacking, the parallel transport method fails to produce satisfactory outcomes. This underscores the significance of considering the similarity of local subspaces.

% \paragraph{\jo{Simpler datasets exhibit more consistent tangent spaces over timesteps.}}
\paragraph{DMs trained on simpler datasets exhibit more consistent tangent spaces over time.} 
% \paragraph{DMs trained on simple datasets have similar tangent spaces across different timesteps.} 
% \paragraph{Tangent basis evolves less if trained on simple dataset}
\label{sec:evolution-t}
% \modify{frequency는 다른데, 과연 그 representation도 많이 다를까? -> tangent space에서의 유사함은 representation의 유사함을 말한다. 이런 직관 설명 내용 추가.}
% In the previous section, we studied the evolution of the tangent space by comparing it across different samples. This time, we compare tangent space across different timesteps.
% In the previous section, we studied the evolution of the tangent space by comparing different samples. This time, we investigate how the tangent space change by comparing it across different timesteps.

% This suggests that the local latent basis that the model pays attention to at each point in time is similar. To help readers understand this, we illustrate in Figure 1b how much of the original feature captured by vector vi is lost when we transport it from one timestep to another, using parallel transport from t = T to 0.1T. As expected, when the local tangent space is similar, the transported vector maintains its existing signal, but it loses this pattern as we move further away from the original timestep. We found this relationship between local tangent space and local semantic subspace to hold true for all datasets. For a more detailed discussion, please refer to the appendix.

\begin{figure}[!t]
    \centering
    \includegraphics[width=0.9\linewidth]{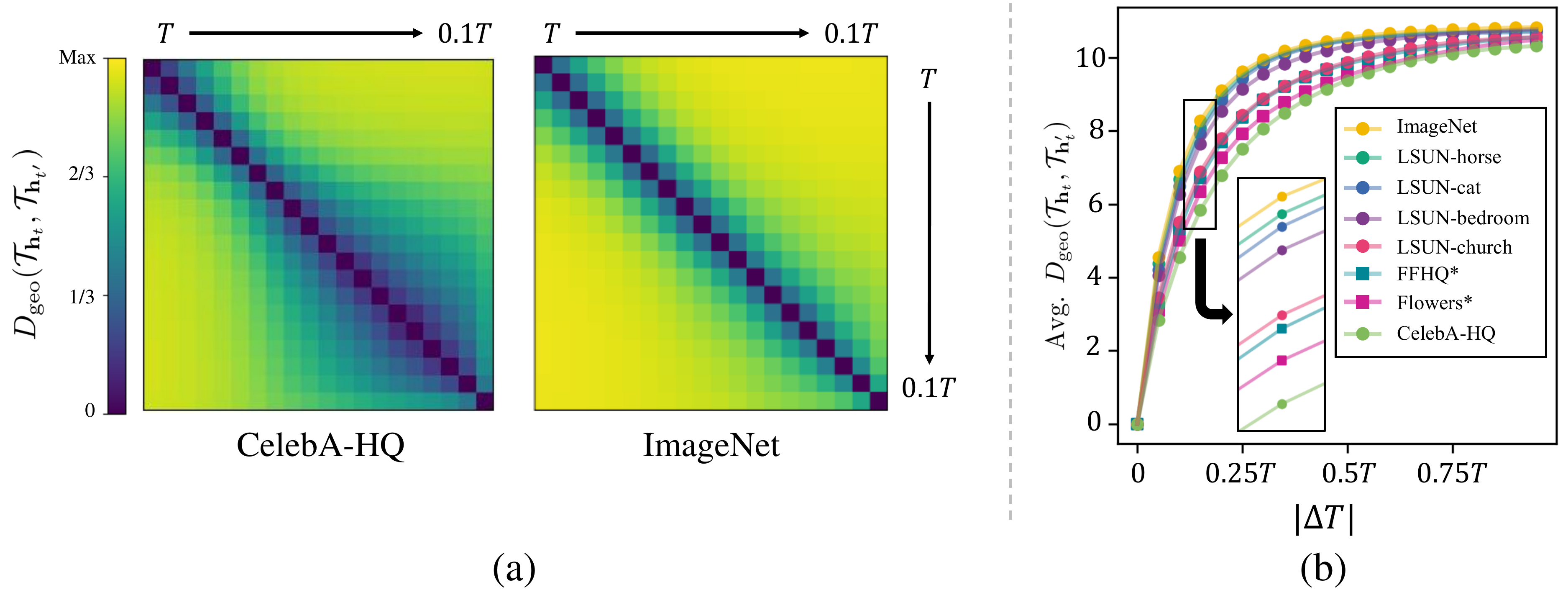}
    \vspace{-0.5em}
    \caption{
    \textbf{\jo{Simpler} datasets lead to \jo{more} similar tangent spaces across diffusion timesteps.}
    % \textbf{Models trained on simple datasets have similar tangent spaces at different timesteps.}
    \jo{(a) Distance matrix visualization of tangent space measured by geodesic metric across various timesteps.}
    %(a) Visualization of the distance matrix of tangent space across various timesteps measured by geodesic metric. 
    % (b) Visualization of the result from parallel transport across timesteps. \yh{${\vv'}_i^{t_a \rightarrow t_b}$ denotes the latent vector transported from $t_a$ to $t_b$. Transported vector significantly deviates from the original vector, as the tangent space grows further apart according to the distance matrix. For visualization purposes, $\vv_i$ \jo{is min-max normalized}.} %undergoes min-max normalization.}
    (b) Average geodesic distance \jo{based on timestep differences, indicating that the complexity of the dataset correlates with greater distances between tangent spaces.}
    %according to the difference of  timestep. It shows that the more complex the dataset, the greater the distance between tangent spaces.
    }
    \vspace{-1em}
    \label{fig:timestep}
\end{figure}

In \fref{fig:timestep} (a), we provide a distance matrix of the tangent spaces across different timesteps, measured by the geodesic metric. 
We observe that the tangent spaces are more similar to each other when a model is trained on CelebA-HQ, compared to ImageNet.
\uh{To verify this trend, we measure the geodesic distances between tangent spaces of different timesteps and plot the average distances of the same difference in timestep in \fref{fig:timestep} (b)}.
% To verify this trend, in \fref{fig:timestep} (c), we measured the average geodesic distance of the tangent space according to the difference of diffusion timesteps.
As expected, we find that DMs trained on datasets, that are generally considered simpler, have similar local tangent spaces over time.

% 앞에서 우리는 DM 이 민감하게 반응하는 signal 의 frequency 가 timestep 에 따라 변화함을 확인했다. 그렇다면, 그 representation 은 어떨까?
% 이를 확인하기 위해, 이미지를 생성해나가는 과정의 각 timestep 에서 만들어지는 local tangent subspace 간 geodesic metric 을 측정했다. 

%%% ICML ver
% We further investigate the coarse-to-fine editing along the generative process from timestep $T$ to $0$. \fref{fig:PSD} (a) shows the example directions $\vv_i$ across different timesteps. At $T$, $\vv_i$ leads to coarse attribute changes in $\vx_0$ by blurry change in $\vx_T$. At $0.25T$, $\vv_i$ edits high-frequency details in both $\vx_0$ and $\vx_t$. \fref{fig:PSD} (b) shows the power spectral density (PSD) of $\vv_i$. We compute the PSD by taking $\vv_1, ..., \vv_{10}$ from 20 samples. The early timesteps contain a larger portion of low frequency than the later timesteps and the later timesteps contain a larger portion of high frequency. This phenomenon agrees with the tendency in the edited images. This results strengthens the common understanding of the timesteps~\cite{kwon2022diffusion,choi2022perception, daras2022multiresolution}.

\begin{figure}[!t]
    \centering
    \vspace{+1em}
    \includegraphics[width=0.9\linewidth]{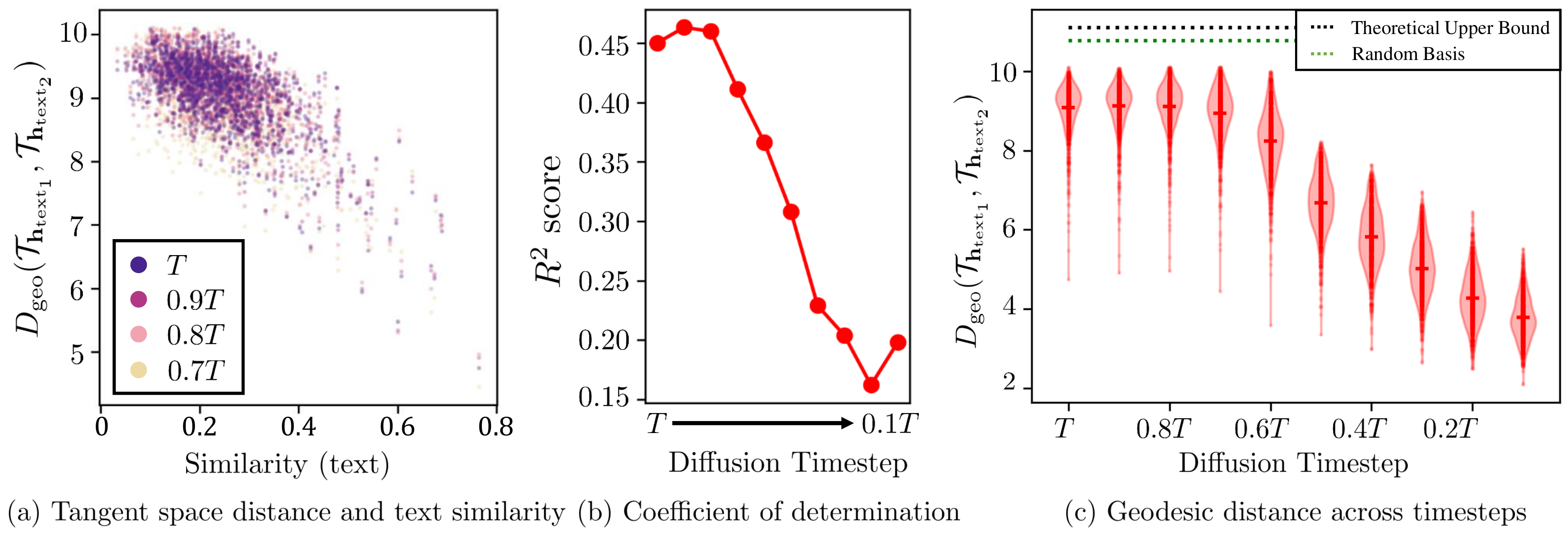}
    % \includegraphics[width=0.9\linewidth]{figure/Homogenity_prompts.pdf}
    % \vspace{-1em}
    \caption{
    \textbf{Similar prompt\jo{s} create similar tangent space\jo{s}, and the \jo{impact of the prompt decreases as the generative process progresses.}} 
    (a) \mingi{The horizontal axis represents the CLIP similarity between two different prompts, \jo{while} the vertical axis represents the geodesic distance in the tangent space from each prompt.} \jo{Different colors represent various diffusion timesteps. A negative relationship is observed between prompt similarity and tangent space distance.}
    %Different diffusion timesteps are denoted as different colors.}
    % \modify{The clip similarity between two different prompts and the geodesic distance in the tangent space created from each respective prompt.}
    % Different colors denote different diffusion timesteps.
    %There is a negative relationship between prompt similarity and tangent space distance.
    %\textbf{The \jo{impact of the prompt as the generative process progresses.}} 
    % \textbf{The influence of the text decreases as the generative process progresses.} 
    (b) The $R^2$ score of the linear regression between clip similarity and geodesic distance of tangent spaces decreases \jo{throughout} the generative process.
    (c) Each point represents the distance between tangent spaces created from different prompts. Until around $t = 0.7T$, the distance between tangent spaces is very large, but it gradually \jo{decreases thereafter.}
    %becomes closer afterward. 
    This indicates that the influence of the prompt on the tangent space diminishes.
    }
    % \vspace{-1em}
    \label{fig:stable_text}
\end{figure}

%\subsection{How the Text-Condition Controls the Geometry of DMs?}
\subsection{\uh{Effect of conditioning prompts on the latent structure}}
% \subsection{\jo{Text-conditioned control of geometry in latent spaces}}
\label{sec:text}
% In this subsection, we aim to find how prompts controls the generative process in geometrical perspective. 
% To demonstrate this, we analyze local semantic subspace and tangent space from randomly sample 50 captions from the MS-coco dataset \cite{lin2014microsoft}. Here we used $\vx{} \sim \mathcal{N}(0, \textbf{I})$ for latent variable.
% To explore this, we randomly sample 50 captions from the MS-coco dataset \cite{lin2014microsoft} and analyse the local semantic subspace and local tangent space extracted for $\vx{} \sim \mathcal{N}(0, \textbf{I})$. 

% 이를 살펴보기 위해, MS-coco dataset \cite{lin2014microsoft} 로부터 50개의 caption 을 random sample 한 뒤, $\vx{} \sim \mathcal{N}(0, \textbf{I})$ 에 대해 추출된 local semantic subspace 와 local tangent space 를 분석했다. 
% 앞에서 말했듯이, local semantic subspace 는 모델이 주목하는 signal 에 대응하고, local tangent space 는 그 signal 들에 대응하는 semantic represenation 에 대응한다. 

In this subsection, we aim to investigate how prompts influence the generative process from a geometrical perspective. 
We randomly sampled 50 captions from the MS-COCO dataset \cite{lin2014microsoft} and used them as text conditions.
% For the latent variable, we utilize $\vx{} \sim \mathcal{N}(0, \textbf{I})$. 
% Our objective is to explore the impact of text-conditioning on the geometry of DMs.

\paragraph{Similar text conditions induce similar tangent spaces.}
% \fref{fig:stable_text} depicts the relationship between text and the tangent space. 
In \fref{fig:stable_text} (a), we observe a negative correlation between the CLIP similarity of texts and the distance between tangent spaces. 
In other words, when provided with similar texts, the tangent spaces are more similar. 
 % model have similar tangent spaces.
% This implies why manipulating the local latent basis vector according to the given text will result in a corresponding edit (\ifref{fig:local_basis_text_vk}, \ifref{fig:local_basis_text})
The linear relationship between the text and the discrepancy of the tangent spaces is particularly strong in the early phase of the generative process as shown by $R^2$ score in \fref{fig:stable_text} (b).
% In \fref{fig:stable_text} (b),
% the $R^2$ score indicates that the linear relationship between the text and local tangent space is particularly strong during the early stages of the generative process. 

% In this subsection, we aim to find how prompts controls the generative process in geometrical perspective. 
% To demonstrate this, we analyze local semantic subspace and tangent space from randomly sample 50 captions from the MS-coco dataset \cite{lin2014microsoft}. Here we used $\vx{} \sim \mathcal{N}(0, \textbf{I})$ for latent variable.
% To explore this, we randomly sample 50 captions from the MS-coco dataset \cite{lin2014microsoft} and analyse the local semantic subspace and local tangent space extracted for $\vx{} \sim \mathcal{N}(0, \textbf{I})$. 

\paragraph{\uh{The generative process depends less on text conditions in later timesteps.}}
% \paragraph{\jo{Text-conditioning effects diminish over generative processes.}}
%\paragraph{The effect of the text becomes weaker, as the generative process progresses.}
\fref{fig:stable_text} (c) illustrates the \uh{distances between} local tangent spaces for given different prompts with respect to the timesteps.
% prompt 에 따른 local tangent space 의 차이가 generative process 가 진행됨에 따라 어떻게 달라지는지를 보여주고 있다.
%%% mingi
% \fref{fig:stable_text} (c) provides the geodesic distance in the tangent space between all texts and the geodesic distance in the semantic subspace between all texts. 
Notably, as the diffusion timestep approaches values below $0.7T$, the distances between the local tangent spaces start to decrease. 
It implies that the variation due to walking along the local tangent basis depends less on the text conditions, i.e., the text less influences the generative process, in later timesteps.
% This signifies that with smaller timesteps, the variation in the local tangent space becomes less dependent on the text, implying a reduced influence of the text. 
% \mingi{
% Furthermore, we also discover that the geodesic distance in the semantic subspace is maximized at $0.7T$. This observation highlights the disparity in the basis at $0.7T$, thus implying why performing edits at around $0.7T$ is well.
% }
It is a possible reason why the correlation between the similarity of prompts and the similarity of tangent spaces reduces over timesteps.

%% file: 5conclusion.tex
\section{Discussion}
\begin{figure}[!t]
    \centering
    \includegraphics[width=1.0\linewidth]{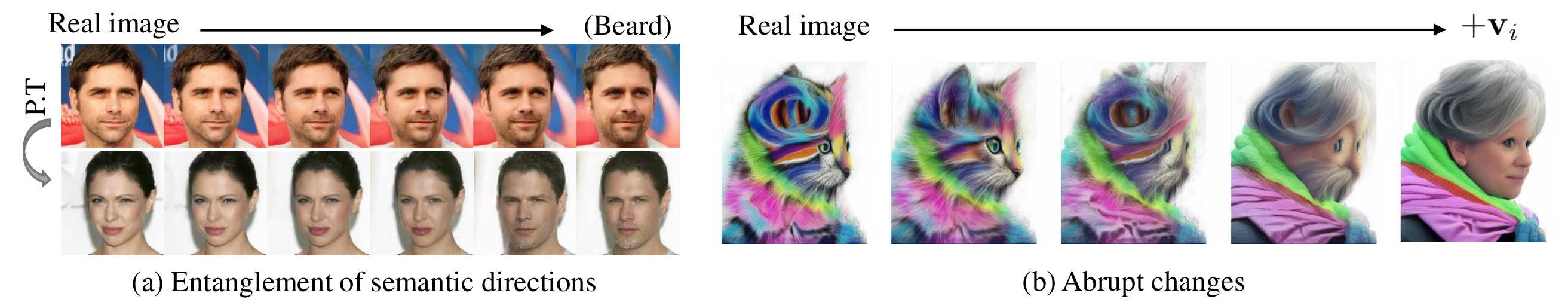}
    \vspace{-1.5em}
    \caption{
    \textbf{Limitations.} (a) Entanglement between attributes due to \jo{dataset biases}
    %the dataset prior. 
    (b) Abrupt changes in Stable Diffusion.}
    \vspace{-1em}
    \label{fig:limitation}
\end{figure}

% \vspace{-0.5em}
% \modify{tangent space가 아닌 latent subspace에서 editing을 했기 때문에 continues하지 않은 것 같다. strength 이야기 넣어도 좋을 듯? 무한히 더하면 이미지 망가진다?}
In this section, we provide additional intuitions and implications. It is interesting that our latent basis usually conveys disentangled attributes even though we do not adopt attribute annotation to enforce disentanglement. We suppose that decomposing the Jacobian of the encoder in the U-Nets naturally yields disentanglement to some extent. 
% It grounds on the linearity of the intermediate feature space $\mathcal{H}$ in the U-Nets \cite{kwon2022diffusion}. 
However, it does not guarantee the perfect disentanglement and some directions are entangled. For example, the editing for beard converts a female subject to a male as shown in \fref{fig:limitation} (a). This kind of entanglement often occurs in other editing methods due to the dataset bias: female faces seldom have beard.

While our method has shown effectiveness in Stable Diffusion, more research is needed to fully validate its potential.
We have observed that some of the discovered latent vector occasionally leads to abrupt changes during the editing process in Stable Diffusion, as depicted in Figure \ref{fig:limitation} (b). 
% In contrast to GANs, where changes tend to occur continuously, we often encounter instances where the transformations appear more like the result of crushing objects, manifesting as discontinuous changes. 
This observation highlights the complex geometry of $\mathcal{X}$ in achieving seamless editing. Exploring this topic in future research is an interesting area to delve into.

\modify{
Our approach is broadly applicable when the feature space in the DM adheres to a Euclidean metric, as demonstrated by \(\mathcal{H}\). This characteristic has been observed in the context of U-Net within \citet{kwon2022diffusion}. It would be intriguing to investigate if other architectural designs, especially those similar to transformer structures as introduced in \cite{peebles2023scalable, tevet2022human}, also exhibit a Euclidean metric.
}

% Although we have shown that our method is also valid to Stable Diffusion, we still need more observation. It discovers less number of latent latent directions and few directions occasionally convey abrupt changes during the editing procedure in Stable Diffusion as shown in \fref{fig:limitation} (b). We suppose that its learned latent space may have a more complex manifold than the image space \cite{arvanitidis2017latent}. Alternatively, the conditional DMs with classifier-free guidance or the cross-attention mechanism may add complexity on the manifold. Our future work includes analyzing the latent directions in the conditions such as text prompts or segmentation labels.

Despite these limitations, our method provides a significant advance in the field of image editing for DMs, and provides a deep understanding of DM through several experiments.

% \vspace{-0.5em}
\section{Conclusion}
\vspace{-0.5em}
We have analyzed the latent space of DMs from a geometrical perspective. 
We used the pullback metric to identify the latent and tangent bases in $\mathcal{X}$ and $\mathcal{H}$. 
The latent basis found by the pullback metric allows editing images by traversal along the basis.
% We can edit images by traversal along the latent directions.
We have observed properties of the bases in two aspects.
% We investigated the discovered structure from two aspects.
% First, we examined how the latent structure changes over time, and 
First, we discovered that 1) the latent bases evolve from low- to high-frequency components; 2) the discrepancy of tangent spaces from different samples increases along the generative process; and 3) DMs trained on simpler datasets exhibit more consistent tangent spaces over timesteps.
Second, we investigated how the latent structure changes based on the text conditions in Stable Diffusion, and discovered that similar prompts make tangent space analogous but its effect becomes weaker over timesteps.
We believe that a better understanding of the geometry of DMs will open up new possibilities for adopting DMs in useful applications.

\vspace{-0.5em}
\section{Acknowledgement}
\vspace{-0.5em}
\jo{This work was supported in part by the Creative-Pioneering Researchers Program through Seoul National University, the National Research Foundation of Korea (NRF) grant (Grant No. 2022R1A2C1006871) (J. J.)}, KIAS Individual Grant [AP087501] via the Center for AI and Natural Sciences at Korea Institute for Advanced Study, and the National Research Foundation of Korea (NRF) grant (RS-2023-00223062).

%% file: 6appendix.tex
\appendix
\onecolumn

\renewcommand{\thetable}{A\arabic{table}}
\renewcommand{\thefigure}{A\arabic{figure}}
\setcounter{figure}{0}
\setcounter{table}{0}
\renewcommand{\theHtable}{A\arabic{table}}
\renewcommand{\theHfigure}{A\arabic{figure}}

\section{Societal Impacts \& Ethics Statements}
\label{appendix:social_impact}
Our research endeavors to unravel the geometric structures of the diffusion model and facilitate high-quality image editing within its framework. While our primary application resides within the creative realm, it is important to acknowledge that image manipulation techniques, such as the one proposed in our method, hold the potential for misuse, including the dissemination of misinformation or potential privacy implications. Therefore, the continuous advancement of technologies aimed at thwarting or identifying manipulations rooted in generative models remains of utmost significance.

\section{Implementation details}
\label{appendix:implementation_detail}

\paragraph{Models and datasets}
We validate our method and provide analyses on various models using the official code and pre-trained checkpoints. The available combinations of the models and the datasets are: 
DDPM~\cite{ho2020denoising} on ImageNet~\cite{deng2009imagenet}, LSUN-church/bedroom/cat/horse~\cite{yu2015lsun}, and CelebA-HQ~\cite{karras2018progressive}; and DDPM trained with {\it P2 weighting}~\cite{choi2022perception} on FFHQ~\cite{karras2019style}, Flowers~\cite{yu2015lsun} and AFHQ~\cite{choi2020stargan}. We also use Stable Diffusion (SD) version 2.1~\cite{rombach2022high} for the text-conditional diffusion model.
% including ImageNet~\cite{deng2009imagenet}, LSUN-church/bedroom/cat/horse~\cite{yu2015lsun}, and CelebA-HQ~\cite{karras2018progressive} for DDPM~\cite{ho2020denoising}; and FFHQ~\cite{karras2019style}, Flowers~\cite{yu2015lsun} and AFHQ~\cite{choi2020stargan} for DDPM trained with {\it P2 weighting}~\cite{choi2022perception}. We also use Stable Diffusion (SD) version 2.1~\cite{rombach2022high} for the text-conditional diffusion model.

For image editing, we use the official codes and pre-trained checkpoints for all baselines and keep the parameters \textit{frozen}. 
For analysis, we compare models with the same diffusion scheduling (linear schedule) and resolutions ($256^2$) to ensure a fair comparison, except Stable Diffusion.

\tref{tab:hyperparameter}1 summarizes various hyperparameter settings in our experiments. Specific details not covered in the main text are discussed in the following paragraphs.

\paragraph{Edit timestep ($t_{edit}$)}
For unconditional DMs, we show the editing results at $t_{edit} \in \{T, 0.8T, 0.6T\}$, while for Stable Diffusion, we show the editing results at $t_{edit} \in \{0.7T, 0.6T\}$. Note that our method allows manipulation at any timestep. 

\paragraph{Inversion step}
We conduct real image editing with DDIM inversion \cite{song2020denoising}. We set the number of steps to $100$ for obtaining the latent variable $\mathbf{x}_T$ and all experiments.
% real image editing 을 위해, DDIM inversion 을 통해 latent variable xT 를 얻었다. \cite{song2020denoising} 이때, 모든 실험에 대해 step 의 개수는 은 둘다 100 으로 고정했다. 

\paragraph{$\vx$-space guidance scale ($\gamma$)}
The value of $\gamma$ determines the magnitude of a single editing step by $\vx$-space guidance. Fortunately, through experimentation, we observed that the value of $\gamma$ does not have a significant impact on image quality unless it is excessively large.

% To obtain the latent code of a given image, we compute the latent code $\mathbf{x}_T$ using DDIM inversion. \cite{song2020denoising} The inversion step hyperparameter refers to the number of DDIM steps used to calculate the latent code. 

\paragraph{Low-rank approximation ($n$)}
We employ a low-rank approximation of the tangent space using $n=50$ for all settings.
% \yh{모든 setting 에서 $n=50$ 을 사용함. pca 를 통해 eigenvalue spectrum 을 그려주면 좋음.}

% In our work, we employ a low-dimensional approximation of the tangent space. Rather than fixing the dimensionality at $n$, we determined to dynamically choose $n$ based on the distribution of eigenvalues. More specifically, we approximated the tangent space with dimensions corresponding to eigenvalues with cumulative density below a given threshold. As such, Table 1 presents the threshold rather than the dimensionality $n$. It worth note that, despite being determined dynamically, the actual values of $n$ has stable for various images. For example, for $t = T, 0.75T, 0.5T, 0.25T$, the values of $n$ were approximately 25, 50, 75, and 100, respectively.

\paragraph{Quality boosting ($t_{boost}$)} While DDIM alone already generates high-quality images, \citet{karras2022elucidating} showed that including stochasticity in the process improves image quality and \citet{kwon2022diffusion} suggest similar technique: adding stochasticity at the end of the generative process. We employ this technique in our experiments on every experiment  after $t=0.2T$, except Stable Diffusion.

\paragraph{Computing resource}
For power-method approximation with $n=50$, it spends about 3-4 minutes on a single NVIDIA RTX 3090 (24GB). As $n$ specifies the number of bases, it can be as small as a user want to use for image editing. Reducing $n$ provides faster runtime, e.g., 10 seconds for $n=3$.
% One can use the power-method approximation with $n<50$ for saving time-consuming when only needing image editing.
% 우리는 NVIDIA 3090 24GB 1대로 모든 실험을 돌렸다. power-method approximation 을 통해 latent basis 를 구할때는, n 에 따라 다르지만, Stable Diffusion 을 포함한 모든 모델이 3~4분 내외로 걸린다. 만약, 이번 work 와 같이 분석이 목적이 아니라 editing 이 목적이면 더 작은 $n<50$ 을 사용할 수 있다. 이 경우에는 n 이 작을수록 걸리는 시간이 줄어든다. 

% \paragraph{Stable Diffusion}
% In order to mitigate the influence of classifier-free guidance, the strength of the guidance, denoted as $w$, was set to zero, utilizing only the text-conditional model. \cite{ho2022classifier} When generating the original Cyberpunk city images, we set the guidance strength as $w = 7.5$. The prompts utilized for the Cyberpunk city images were ``Cyberpunk city" and for the Van Gogh paintings, the prompt used was ``painting of Van Gogh." Through the process of DDIM inversion, latent codes $\mathbf{x}_T$, were generated given the appropriate prompts for each image, with the guidance strength also set to zero (i.e., $\text{guidance scale} = 1$ in the code).

\begin{table}[t]
\caption{Hyper-parameter settings.}
\label{tab:setting}
\begin{center}
\begin{small}
% \begin{sc}
\begin{tabular}{lcccccc}
\toprule
model & $t_{edit}$ & inversion step & $\gamma$ & $n$ & $t_{boost}$ \\
\midrule
Stable Diffusion    & $0.7T$ & 100 & 1     & 50 & $\times$  \\
                    & $0.6T$ & 100 & 2     & 50 & $\times$  \\
Unconditional DMs   & $T$    & 100 & 0.5   & 50 & $0.2T$    \\
                    & $0.8T$ & 100 & 1     & 50 & $0.2T$    \\
                    & $0.6T$ & 100 & 4     & 50 & $0.2T$    \\
\bottomrule
\end{tabular}
% \end{sc}
\end{small}
\end{center}
\vskip -0.1in
\end{table}
\label{tab:hyperparameter}
\section{Ablation study}

In this section, we validate our method with ablation study. 

\begin{figure}[!t]
    \centering
    \includegraphics[width=0.9\linewidth]{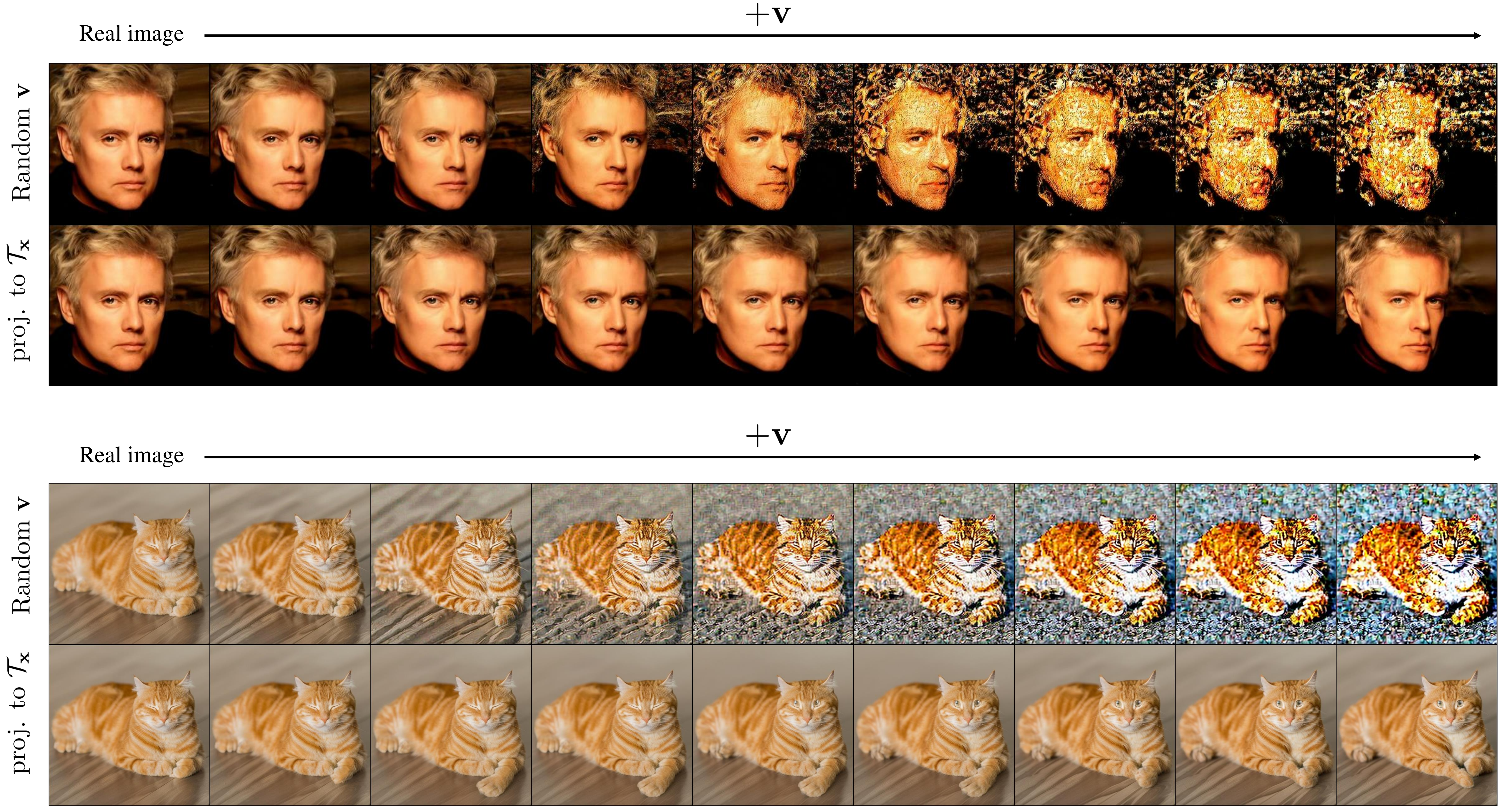}
    \caption{
    \textbf{Importance of the discovered latent directions.} Random direction experiments with CelebA-HQ pre-trained model (top) and Stable Diffusion (bottom).
    % 위 3개 row 는 CelebA-HQ 에서의 결과, 밑은 Stable Diffusion 에서의 결과를 보여주고 있다. 
    %The image on the far left represents the reconstructed original image, while the subsequent images demonstrate the interpretable edits that have been made to it. 
    % {The leftmost image represents the reconstruction of real image. }
    \yh{Adding random directions instead of latent directions severely distorts the resulting images.}
    %As we move towards the right, we impose stronger editing. 
    %When we perform edits along a random direction, the generated image does not change semantically meaningfully.
    {When we perform edits along the projection onto the latent subspace $\mathcal{T}_{\mathbf{x}}$, the generated image presents a semantically meaningful transformation.
    }
    % Projection onto a space orthogonal to the subspace further supports it by showing less semantic change and consistent image distortion.
    }
    \label{fig:random_v}
\end{figure}

\paragraph{Random $\vv$} 
To demonstrate the meaningfulness of the latent basis found by our method, we qualitatively compare its effect to na\"ive baseline: random directions.
% To demonstrate that the latent basis we obtained is meaningful, we provide experiments using random directions.
The first row in \fref{fig:random_v} shows that manipulating the images with a random vector `$\vv$' does not result in semantic editing but rather degrades images.
The second row shows the results of projecting the random `$\vv$' onto our obtained latent subspace. The projected results exhibit semantic manipulation such as pose changes without image distortion.
It indicates that the found latent subspace captures semantics in the latent space effectively.

% To verify the quality of the latent basis we obtained, we conducted a comparison with a random direction. First, as observed in \fref{fig:random_v}, random $\vv$ does not effectively allow for semantic editing. Furthermore, projecting the random $\vv$ onto the latent subspace reveals significant semantic manipulation. This implies that the latent subspace we discovered effectively captures the local semantic information of the latent space.

\begin{figure}[!t]
    \centering
    \includegraphics[width=1.0\linewidth]{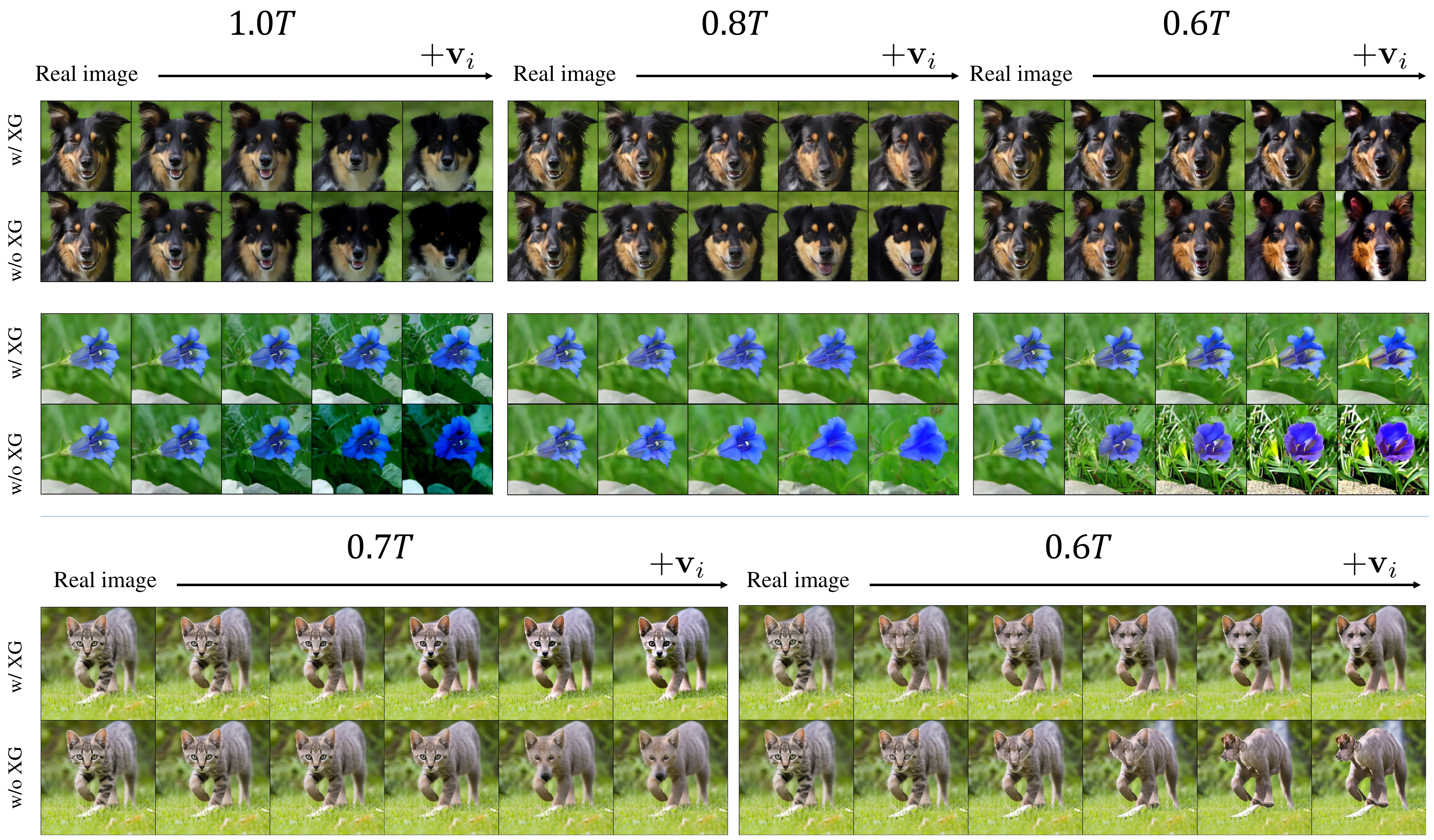}
    \caption{
    \textbf{Importance of the $\vx$-space guidance.} 
    $\vx$-space guidance experiments with AFHQ (top), Flowers pre-trained model (middle), and Stable Diffusion (bottom).
    $\vx$-space guidance helps achieve qualitatively similar editing while preserving the content of the original image.}
    \label{fig:x_space_guidance}
\end{figure}

\paragraph{$\vx$-space guidance}
\label{appendixsec:ablation_x_guidance}
\fref{fig:x_space_guidance} demonstrates the the effectiveness of $\vx$-space guidance compared to a straightforward alternative: simple addition. First, $\vx$-space guidance produces higher quality images with similar meaning. Especially Stable Diffusion apparently benefits from $\vx$-space guidance regarding smoothness of the editing strength and artifacts. The difference is more significant at $t=0.6T$. Note that the meaning of the same directions may slightly differ between the two settings due to non-linearity of the U-Net.
% To investigate the effectiveness of $\vx$-space guidance, we qualitatively compare it to the na\"ive approach of direct addition. First, as shown in \fref{fig:x_space_guidance}, overall, $\vx$-space guidance allows qualitatively similar manipulations while enhancing their quality. Specifically, we observe that when using $\vx$-space guidance, artifacts are reduced to a greater extent in Stable Diffusion compared to the unconditional model, and the reduction is more significant at $t=0.6T$ compared to $t=T$. However, it is worth noting that the qualitative direction of the manipulations may slightly differ.

Currently, we do not have a deeper understanding of the underlying principles of $\vx$-space guidance. Exploring the reasons behind its ability to improve manipulation quality would be an interesting direction for future work.

% \clearpage

\begin{figure}[!t]
    \centering
    \includegraphics[width=0.95\linewidth]{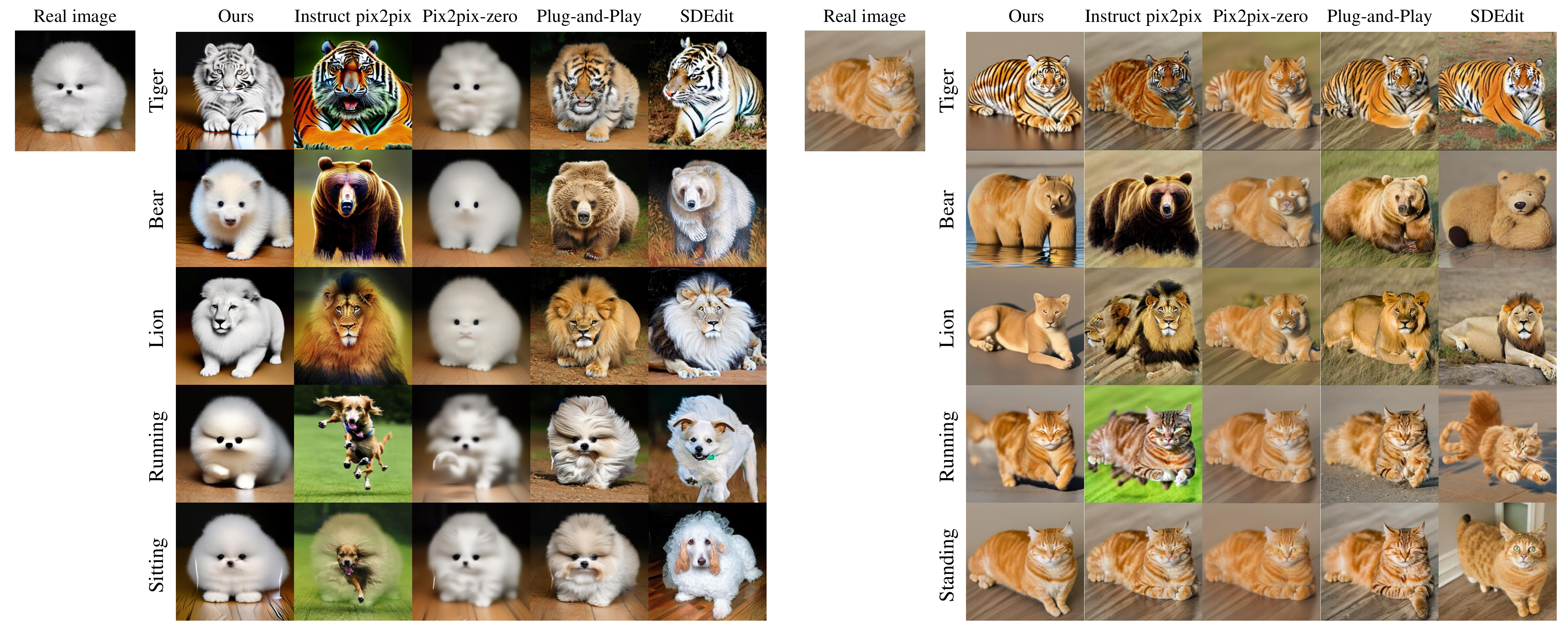}
    \caption{
    \textbf{Comparison with various image editing methods.}
    Our approach empowers image editing that aligns seamlessly with text conditions while upholding object identity. In contrast, alternative methods exhibit deficiencies such as: inadequate preservation of object structure (Instruct pix2pix), inefficacious manipulation (Pix2pix-zero), or challenges in maintaining identity fidelity (Plug-and-Play, SDEdit).
    }
    %\caption{\textbf{Comparison with different image editing methods.} Our method enables image editing that aligned with text conditions while preserving object identity. In contrast, other methods have shortcomings like: not preserving object structure (Instruct pix2pix), failing to manipulate effectively (Pix2Pix-zero), or having issues with less preserved identity (Plug-and-Play).}
    % \vspace{0.2em}
    \label{fig:comparison}
\end{figure}

\section{Comparative experiment to other state-of-the-art (SoTA) editing methods}
\label{appendix:comparisons}
We conduct qualitative comparisons with text-guided image editing methods. Our SoTA baseline methods include: (i) SDEdit \cite{meng2021sdedit}, (ii) Pix2Pix-zero \cite{parmar2023zero}, (iii) PnP \cite{tumanyan2022plug}, and (iv) Instruct Pix2Pix \cite{brooks2023instructpix2pix}. All comparisons were performed using the official code.
Please refer to \fref{fig:comparison} for the qualitative results.

We also compare the time complexity of each method. For a fair comparison, we only identify the first singular vector $\mathbf{v}_1$, i.e., $n=1$, and set the number of DDIM steps to 50. All experiments were conducted on an Nvidia RTX 3090. The runtime for each method is summarized in \tref{tab:comparison_time}.

The computation cost of our method remains comparable to other approaches, although the Jacobian approximation takes around 2.5 seconds for $n=1$. This is because we only need to identify the latent basis vector once at a specific timestep. Furthermore, our approach does not require additional preprocessing steps like generating 100 prompts with GPT and obtaining embedding vectors (as in Pix2Pix-zero), or storing feature vectors, queries, and key values (as in PnP). Our method also does not require finetuning (as in Instruct Pix2Pix). This leads to a significantly reduced total editing process time in comparison to other methods.

\begin{table}[t]
\caption{\modify{\textbf{Comparisons of the time complexity of state-of-the-art editing methods} 
% We used five attribution methods, including a random baseline, to compare the rankings. 
% A stronger correlation between the rankings signifies greater consistency in the results, irrespective of whether the model undergoes retraining or not.
}}
% \vskip 0.15in
\begin{center}
\begin{small}
\begin{tabular}{c|c|c}
\toprule
 Image Edit Method & Running time & Preprocessing \\
\midrule
Ours             &     11 sec    & N/A            \\
SDEdit           &      4 sec    & N/A            \\
Pix2Pix-zero     &     25 sec    & 4 min          \\
PnP              &     10 sec    & 40 sec         \\
Instruct Pix2Pix &     11 sec    & N/A            \\
\bottomrule
\end{tabular}
\end{small}
\end{center}
% \vskip -0.1in
\label{tab:comparison_time}
\end{table}

\section{More Discussions}

% \begin{wrapfigure}{!r}{6cm}
%     \vspace{-1.5em}
%     \includegraphics[width=6cm]{figure/parallel_transport_x_theta.pdf}
%     \vspace{-1.5em}
%     \caption{\textbf{Parallel transport between similar tangent spaces creates similar latent directions.}
%     The horizontal axis represents the geodesic distance between tangent spaces from different timesteps, while the vertical axis represents the angle between the original latent direction and transported latent direction. 
%     Different colors represent various datasets. 
%     A positive relationship is observed between tangent space distance and the distortion induced by parallel transport.}
%     \label{fig:parallel_transport_x_theta}
%     \vspace{-2.5em}
% \end{wrapfigure} 

\begin{figure}[!t]
\centering
    \includegraphics[width=0.95\linewidth]{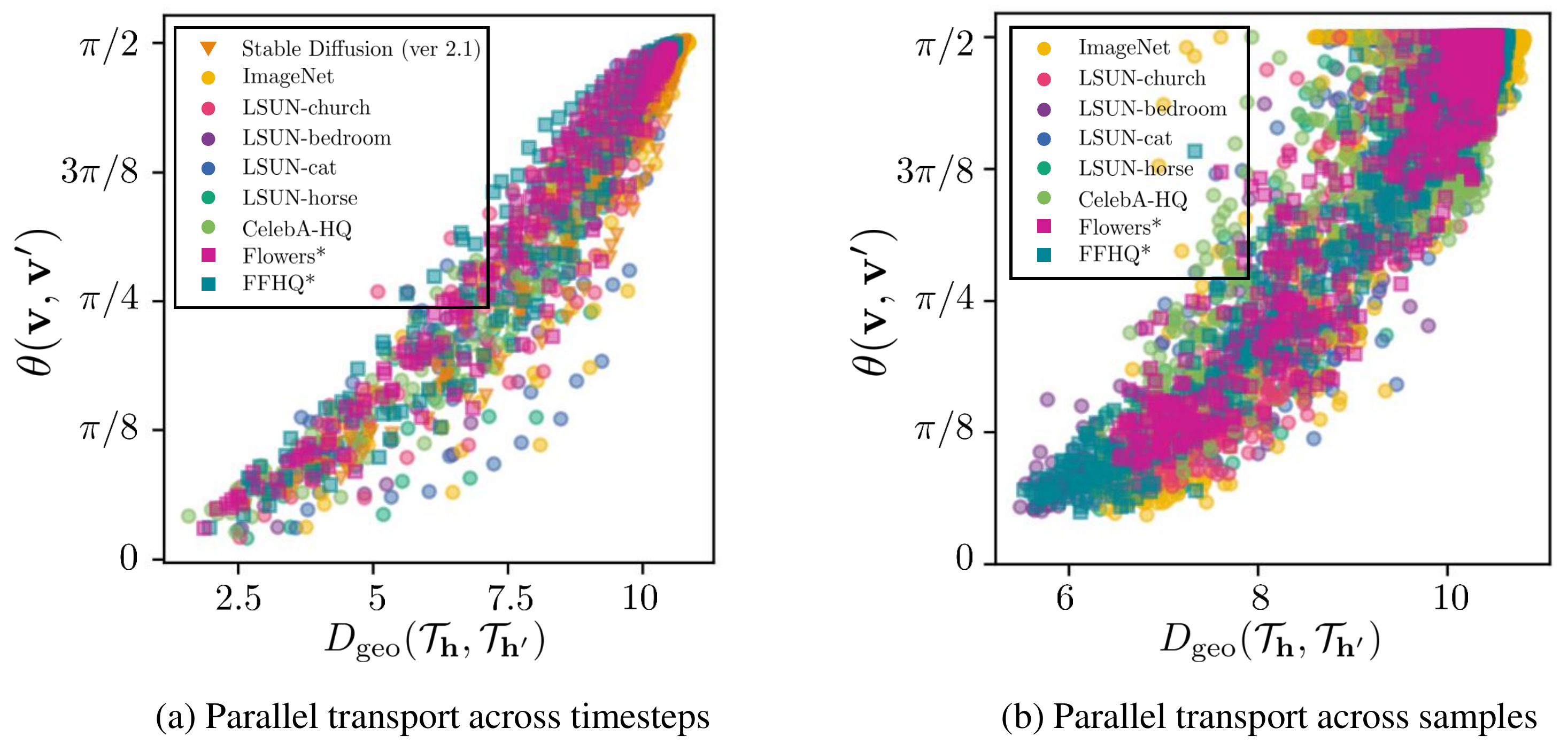}
    \vspace{-0.5em}
    \caption{
    \textbf{Parallel transport between similar tangent spaces creates similar latent directions.} 
    The horizontal axis represents the geodesic distance between tangent spaces from (a) different timesteps (b) different samples at $t \in \{T, 0.9T, \cdots, 0.1T\}$. The vertical axis represents the angle between the original latent direction and transported latent direction. 
    Different colors represent various datasets. 
    A positive relationship is observed between tangent space distance and the distortion induced by parallel transport.
    }
    \label{fig:parallel_transport_x_theta}
    % \vspace{-1em}
\end{figure}

\paragraph{Why do we measure the geodesic distance of the tangent spaces instead of the latent subspaces?}
\label{appendixsec:relationship_tangent_latent}
The geodesic distance on the Grassmannian manifold between two subspaces is defined as the $l_{2}$-norm of principal angles. 
To define angles between different vector spaces, an inner product needs to be defined. In our work, we define the inner product in $\mathcal{T}_\vx$ using the pullback metric. The issue is that the pullback metric is locally defined for each latent subspace $\mathcal{T}_\vx$ (\eref{eq:pullback}). Therefore, measuring angles between distant latent subspaces becomes challenging. On the other hand, \ehspace{} follows the assumption of the Euclidean metric. Consequently, even for distant tangent spaces, angles can be easily computed using the dot product.
In this regard, we measure the similarity between latent subspaces by exploiting the geodesic distance of their corresponding tangent spaces.
\yh{Furthermore, when compared to \exspace{}, \ehspace{} offers the advantage of being a semantic space, making it more suitable for measuring semantic similarity.}

% In our work, much of the analysis consists of measuring the geodesic distance between different tangent spaces, i.e., $D_{\text{geo}}(\mathcal{T}_\vh, \mathcal{T}_{\vh'})$. 
% This is because the geodesic distance measure the principle angle, based on the dot product. (See \aref{appendixsec:algorithm}). This makes sense in \ehspace{}, where the Euclidean metric is assumed, but not in \exspace{}.

\begin{figure}[!t]
\centering
    \includegraphics[width=0.95\linewidth]{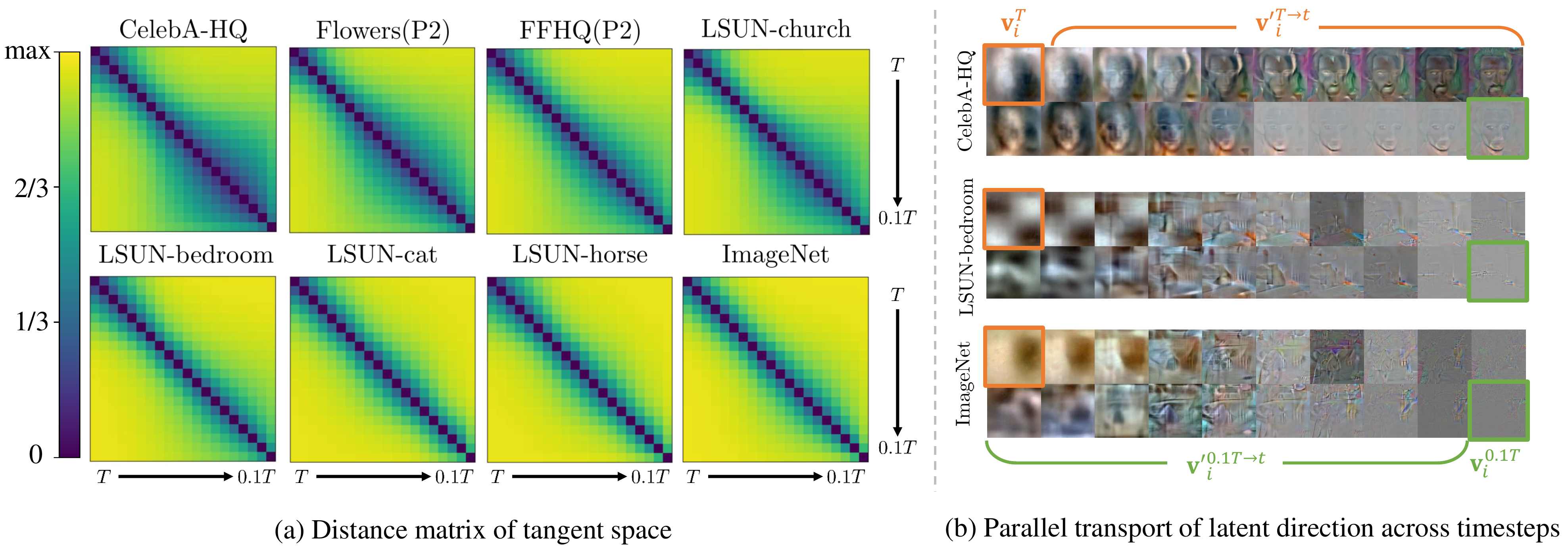}
    \vspace{-0.5em}
    \caption{
    \textbf{More examples of tangent spaces across diffusion timesteps.} 
    (a) Distance matrix visualization of tangent space measured by geodesic metric across various timesteps.
    (b) Visualization of the result from parallel transport across timesteps.
    \yh{${\vv'}_i^{t_a \rightarrow t_b}$ denotes the latent vector transported from $t_a$ to $t_b$.
    Transported vector significantly deviates from the original vector, as the tangent space grows further apart according to the distance matrix. 
    For visualization purposes, $\vv_i$ \jo{is min-max normalized}.}}
    \label{fig:evolution-t-appendix}
    % \vspace{-1em}
\end{figure}

\paragraph{Similar tangent space implies similar latent subspace}
% \paragraph{What does it imply if the tangent space is similar?}
% In \fref{fig:parallel_transport_x_theta}, 서로 다른 timestep 에서 구한 tangent space 간 geodesic distance 와 그 둘 사이에서 parallel transport 한 latent direction 간 각도를 구한 것이다. 
% It is evident that as the geodesic distance decreases, the amount of distortion during parallel transport also reduces.
% \yh{이는 tangent space 가 유사하면 latent subspace 도 유사함을 의미한다.}
In \fref{fig:parallel_transport_x_theta}, we calculated the geodesic distance of tangent spaces obtained at different timesteps (or different samples at same the timestep) and the angle between the original latent direction and parallel transported direction between them.
It is evident that as the geodesic distance decreases, the amount of distortion during parallel transport also reduces.
% For parallel transport across different timesteps, the result is visually illustrated in \fref{fig:evolution-t-appendix}.

\modify{
Notice that the similarity between tangent spaces implies \uh{consistency of latent basis across timesteps}. %a consistent latent basis at each timestep.
In \fref{fig:evolution-t-appendix} (b), we parallel transport the latent vector $\vv_i$ to various tangent spaces and visualize the outcomes.
As expected, when the tangent spaces are similar, the transported vector ${\vv'}_i$ retains the original signal. On the other hand, as we move to more distant timesteps, where the tangent space is farther apart, ${\vv'}_i$ deviates from the original signal.
}

\section{Algorithms}
\label{appendixsec:algorithm}

In this section, for reproducibility purposes, we provide the code for two important algorithms. The code is implemented using PyTorch \cite{paszke2017automatic}.

\paragraph{Jacobian subspace iteration}
The diffusion model has dimensions that are too large in both \exspace{} and \ehspace{}, making the computation of the Jacobian infeasible. To overcome this challenge, we attempt the Jacobian subspace iteration algorithm to approximate the singular value of the Jacobian, as proposed in \cite{haas2023discovering}. For details, please refer to \citet{haas2023discovering}.

\begin{lstlisting}[language=Python, caption=Python example, caption={\textbf{Jacobian subspace iteration}}, captionpos=b]
import torch # >= ver 2.0

def local_encoder_pullback(
        x, t, get_h, n=50, chunk_size=25, min_iter=10, max_iter=100, convergence_threshold=1e-4,
    ):
    '''
    Args
        - x : tensor ; latent variable
        - t : tensor ; diffusion timestep
        - get_h : function ; return h given x, t
        - n ; low-rank approximation dimension
        - chunk_size ; To avoid OOM error
        - min_iter (max_iter) ; minimum (maximum) number of iteration
        - convergence_threshold ; to check convergence of power-method
    '''
    # set number of chunk to avoid OOM
    num_chunk = n // chunk_size + 1

    # get dimensions of x space and h space
    h_shape = get_h(x, t).shape
    c_i, w_i, h_i = x.size(1), x.size(2), x.size(3)
    c_o, w_o, h_o = h_shape[1], h_shape[2], h_shape[3]

    # power-method
    a = torch.tensor(0., device=x.device, dtype=x.dtype)
    vT = torch.randn(c_i*w_i*h_i, n, device=x.device)
    vT, _ = torch.linalg.qr(vT)
    v = vT.T
    v = v.view(-1, c_i, w_i, h_i)

    for i in range(max_iter):
        v = v.to(device=x.device, dtype=x.dtype)
        v_prev = v.detach().cpu().clone()
        
        time_s = time.time()
        u = []
        v_buffer = list(v.chunk(num_chunk))
        for vi in v_buffer:
            g = lambda a : get_h(x + a*vi, t=t)
            ui = torch.func.jacfwd(
                g, argnums=0, has_aux=False, randomness='error'
            )(a)
            u.append(ui.detach().cpu().clone())
        u = torch.cat(u, dim=0)
        u = u.to(x.device, x.dtype)

        g = lambda x : torch.einsum(
            'b c w h, i c w h -> b', u, get_h(x, t=t)
        )
        v_ = torch.autograd.functional.jacobian(g, x)
        v_ = v_.view(-1, c_i*w_i*h_i)

        _, s, v = torch.linalg.svd(v_, full_matrices=False)
        v = v.view(-1, c_i, w_i, h_i)
        u = u.view(-1, c_o, w_o, h_o)
        
        convergence = torch.dist(v_prev, v.detach().cpu()).item()
        if torch.allclose(v_prev, v.detach().cpu(), atol=convergence_threshold) and (i > min_iter):
            break

    # reshape as a x space, h space vector
    u, s, vT = u.reshape(-1, c_o*w_o*h_o).T.detach(), s.sqrt().detach(), v.reshape(-1, c_i*w_i*h_i).detach()
    return u, s, vT
\end{lstlisting}

\paragraph{Geodesic metric}
For a detailed discussion on the geodesic metric, please refer to \citet{choi2021not} for more information.
% Geodesic metric 에 대한 구체적인 논의는 \citet{} 를 참고하라. 

\begin{lstlisting}[language=Python, caption={\textbf{Geodesic metric}}]
import torch

def geodesic_metric(U1, U2):
    _, S, _ = torch.linalg.svd(U1.T @ U2)
    th = torch.acos(S)
    return th.norm()
\end{lstlisting}

% power-method approximation
% \begin{algorithm}[ht!]
% \caption{Feature Direction}
% \label{alg:local_basis}
% \begin{algorithmic}[1]
% \REQUIRE {latent variable $\mathbf{x}$, timestep $t$, U-Net encoder $f : \mathcal{X} \times T \rightarrow \mathcal{H}$, Feature direction index $i$}
% \STATE {$J$             = Jacobian($f(\cdot, t)$)($\mathbf{x}$)}
% \STATE $U, S, V^{\tran}$ = SingularValueDecomposition($J$) 
% \STATE {$\mathbf{v}_i, \mathbf{u}_i$ = $V^{\tran}$[$i$, :], $U$[:, $i$]}
% \STATE {{\bfseries Return} $\mathbf{v}_i, \mathbf{u}_i$}
% \end{algorithmic}
% \end{algorithm}

%%%%%%%%%%%%%%%%%%%%%%%%%%%%%%%%%%%
% below is the additional figures %
%%%%%%%%%%%%%%%%%%%%%%%%%%%%%%%%%%%
\clearpage

\section{Additional results}
\label{appendix:additional_results}
\subsection{Latent basis}
\label{appendix:local}
\begin{figure}[!h]
    \centering
    \includegraphics[width=1.0\linewidth]{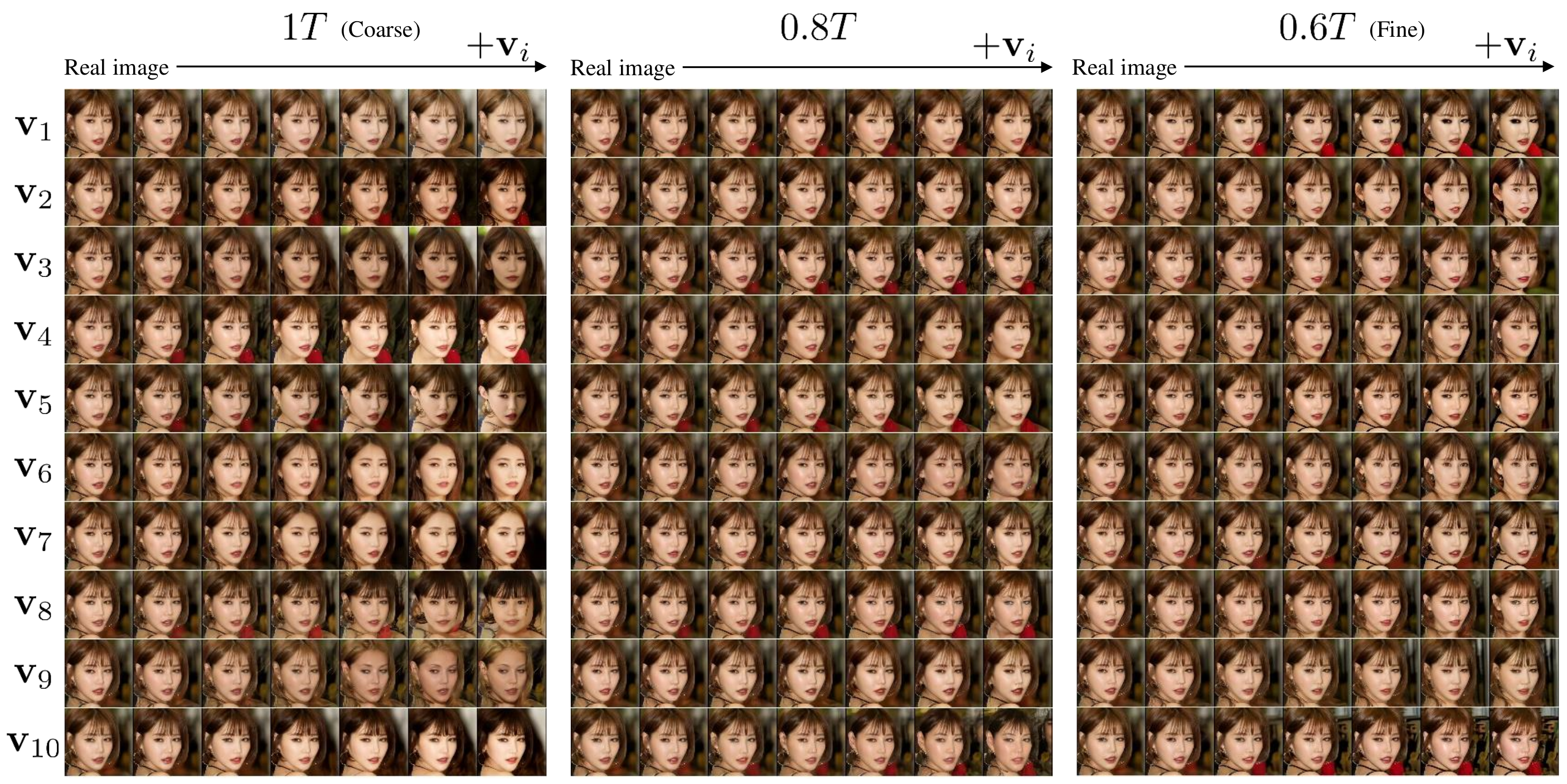}
    \vspace{-2em}
    \caption{
    \textbf{More examples of image editing using the latent basis in FFHQ.} The editing result using ten $\vv_i$' in FFHQ.  Each column represents edits made at different diffusion timesteps (0.6$T$, 0.8$T$, and 1$T$). Editing at timestep 1$T$ yields coarse changes. On the other hand, editing at timestep 0.6$T$ leads to fine changes. Please zoom in for the best view.}
    \label{fig:ffhq_vis}
\end{figure}

\begin{figure}[!h]
    \centering
    \includegraphics[width=1.0\linewidth]{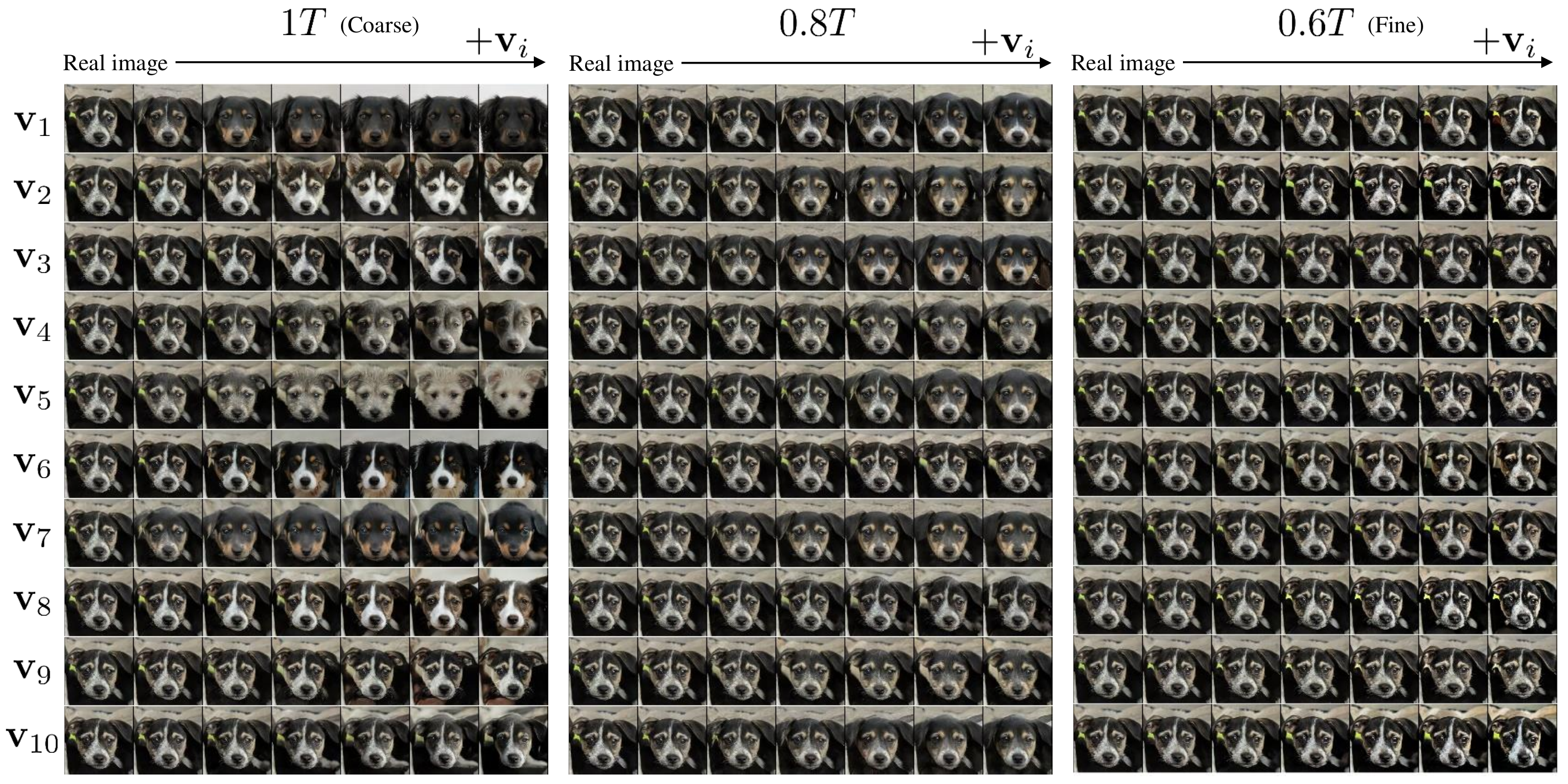}
    \caption{
    \textbf{More examples of image editing using the latent basis in AFHQ.} The editing result using ten $\vv_i$' in AFHQ.  Each column represents edits made at different diffusion timesteps (0.6$T$, 0.8$T$, and 1$T$). Editing at timestep 1$T$ yields coarse changes. On the other hand, editing at timestep 0.6$T$ leads to fine changes. Please zoom in for the best view.}
    \label{fig:afhq_vis}
\end{figure}

\begin{figure}[!h]
    \centering
    \includegraphics[width=1.0\linewidth]{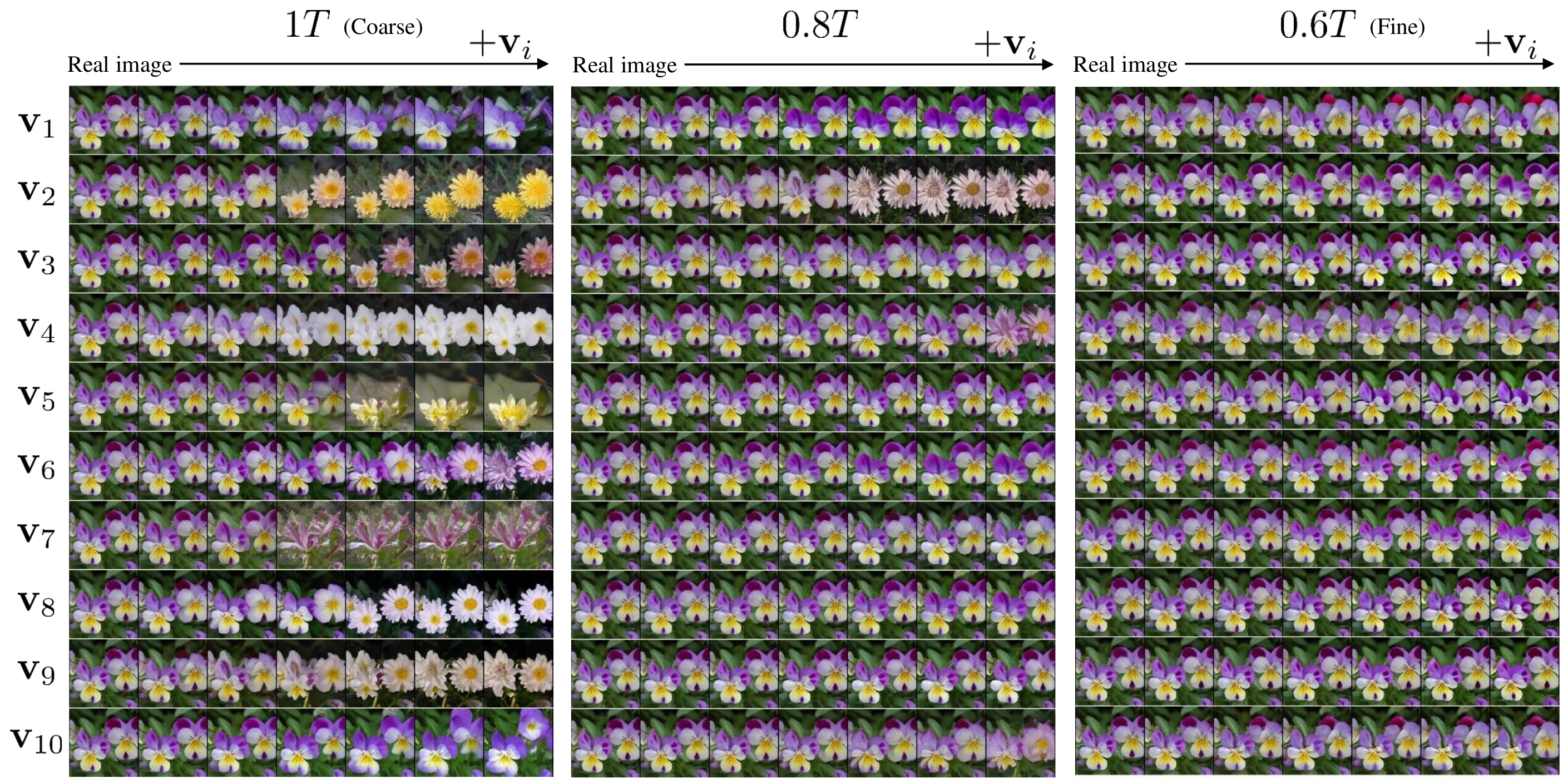}
    \caption{
    \textbf{More examples of image editing using the latent basis in Flowers.} The editing result using ten $\vv_i$' in Flowers.  Each column represents edits made at different diffusion timesteps (0.6$T$, 0.8$T$, and 1$T$). Editing at timestep 1$T$ yields coarse changes. On the other hand, editing at timestep 0.6$T$ leads to fine changes. Please zoom in for the best view.}
    \label{fig:flowers_vis}
\end{figure}

\begin{figure}[!h]
    \centering
    \includegraphics[width=1.0\linewidth]{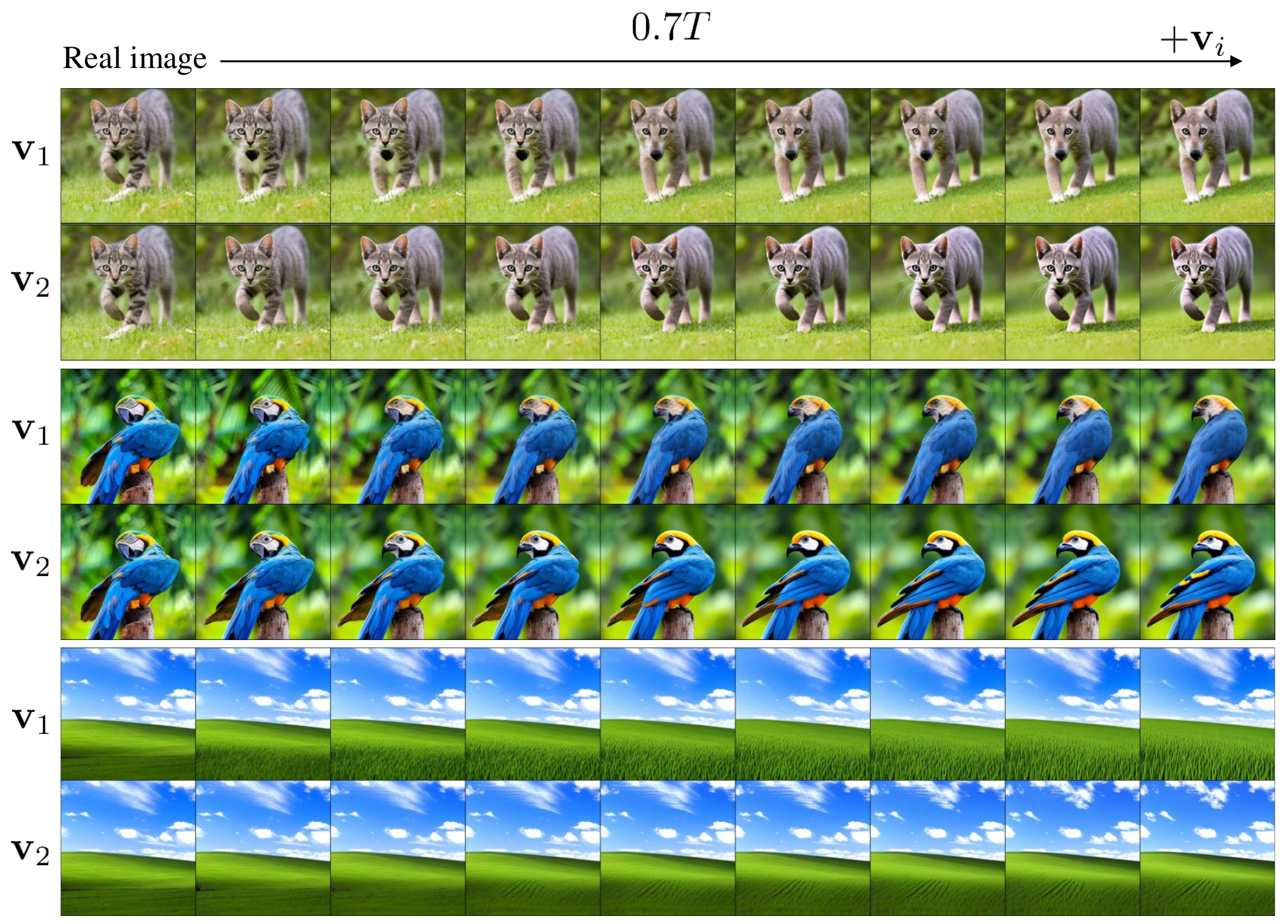}
    \caption{
    \textbf{More examples of image editing using the latent basis with Stable Diffusion.} The editing result using $\vv_i$' in Stable diffusion at 0.7$T$.}
    \label{fig:stable_7T}
\end{figure}

\begin{figure}[!h]
    \centering
    \includegraphics[width=1.0\linewidth]{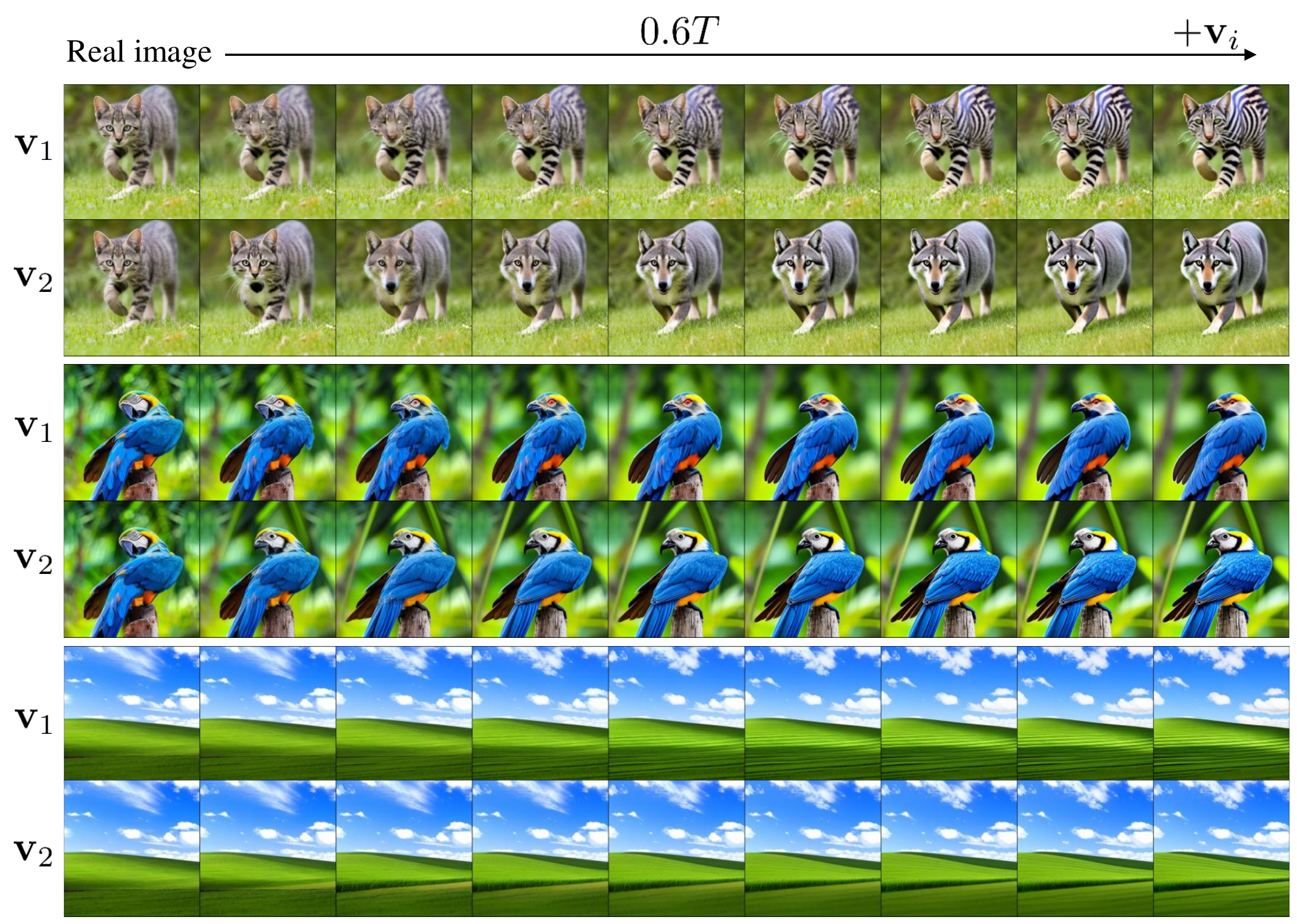}
    \caption{
    \textbf{More examples of image editing using the latent basis with Stable Diffusion.} The editing result using $\vv_i$' in Stable diffusion at 0.6$T$.}
    \label{fig:stable_6T}
\end{figure}

\paragraph{Unconditional latent basis}
We provide more examples of image editing using the latent basis. Figure~\ref{fig:ffhq_vis}, \ref{fig:afhq_vis}, \ref{fig:flowers_vis}, \ref{fig:stable_7T} and \ref{fig:stable_6T} show that every latent basis produces different results and editing at timestep $T$ yields coarse changes while 0.6$T$ leads to fine changes. Stable Diffusion shows a similar trend; 0.7$T$ yields coarse changes while 0.6$T$ leads to fine changes. The results of $T$ in Stable Diffusion will be covered in the \sref{appen:more_discussion}. Please zoom in for the best view.

\clearpage

\begin{figure}[!h]
    \centering
    \includegraphics[width=1.0\linewidth]{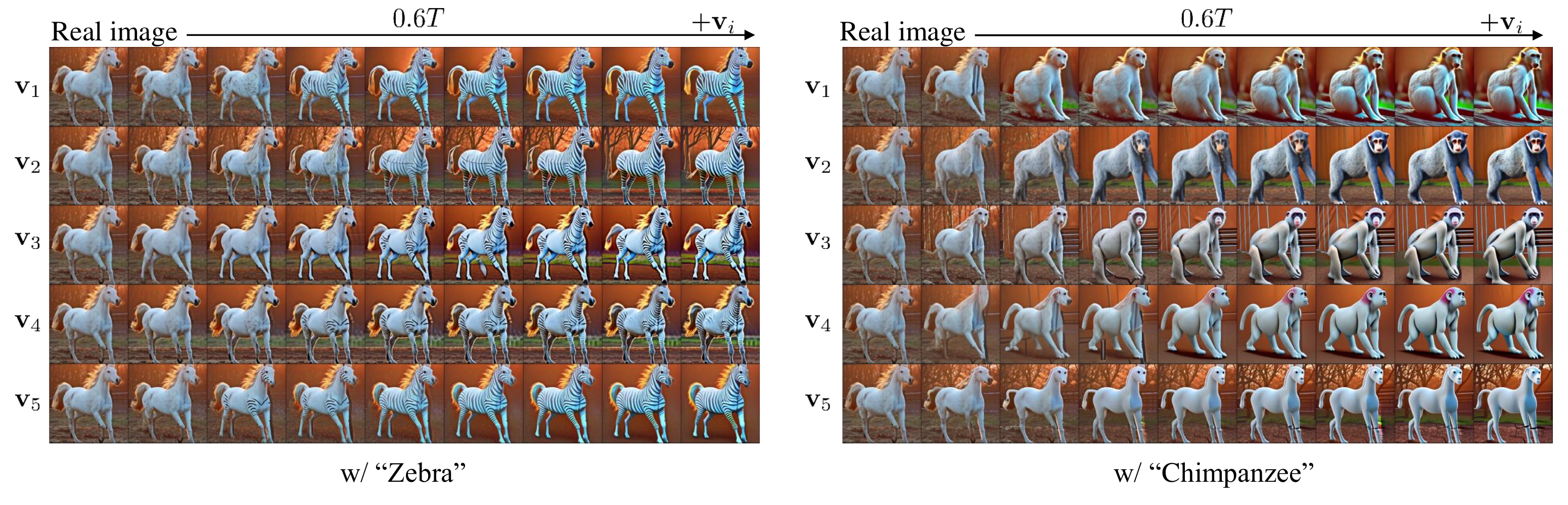}
    \vspace{-2em}
    \caption{
    \textbf{More examples of image editing using top-5 latent basis vectors when given the prompt.} Notably, Changes to ``zebras'', which are clear, all show similar results, but ``chimpanzees'' show different results. Nevertheless, it is clear that it is all related to ``chimpanzees''.}
    \label{fig:stable_various_vis}
\end{figure}

\paragraph{latent basis with given prompt}
%As shown in \fref{fig:stable_various_vis}, using the prompt as a condition, the entire latent basis captures prompt-related attributes. 
As shown in \fref{fig:stable_various_vis}, when we condition a specific prompt, such as ``Zebra'' or ``Chimpanzee'', the entire latent basis corresponds to the prompt-related attributes.
Notably, Changes to ``zebra'', which are clear, all show similar results, but ``chimpanzee'' show different results. Nevertheless, it is clear that they are all related to ``chimpanzee''.

% \subsection{}
\subsection{Image editing using latent basis vectors discovered with various prompts}

\begin{figure}[!h]
    \centering
    \includegraphics[width=1.0\linewidth]{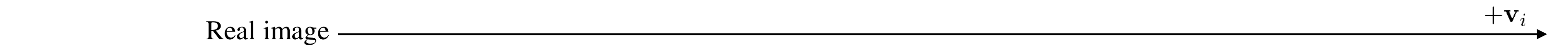}
    \includegraphics[width=1.0\linewidth]{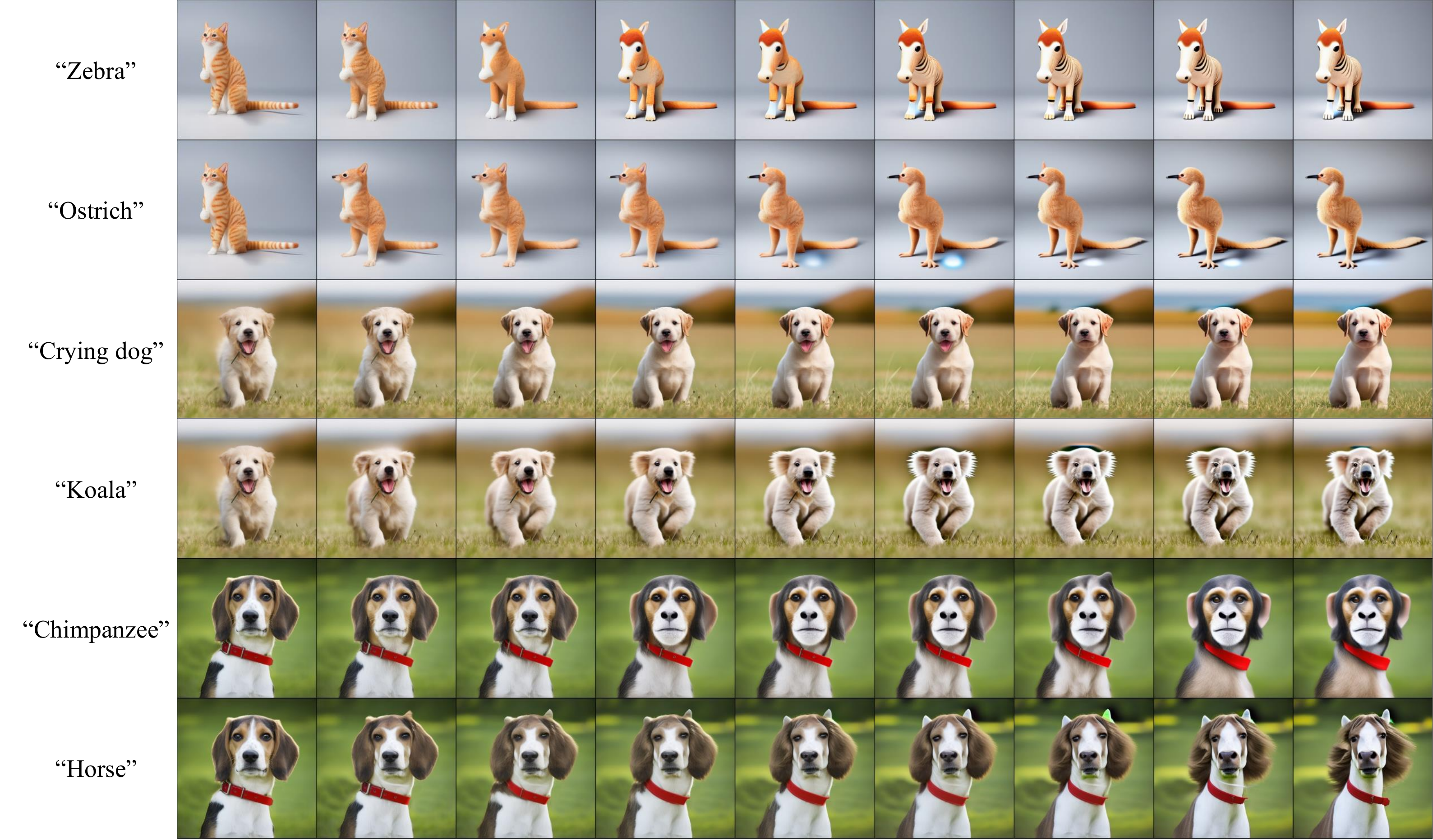}
    \includegraphics[width=1.0\linewidth]{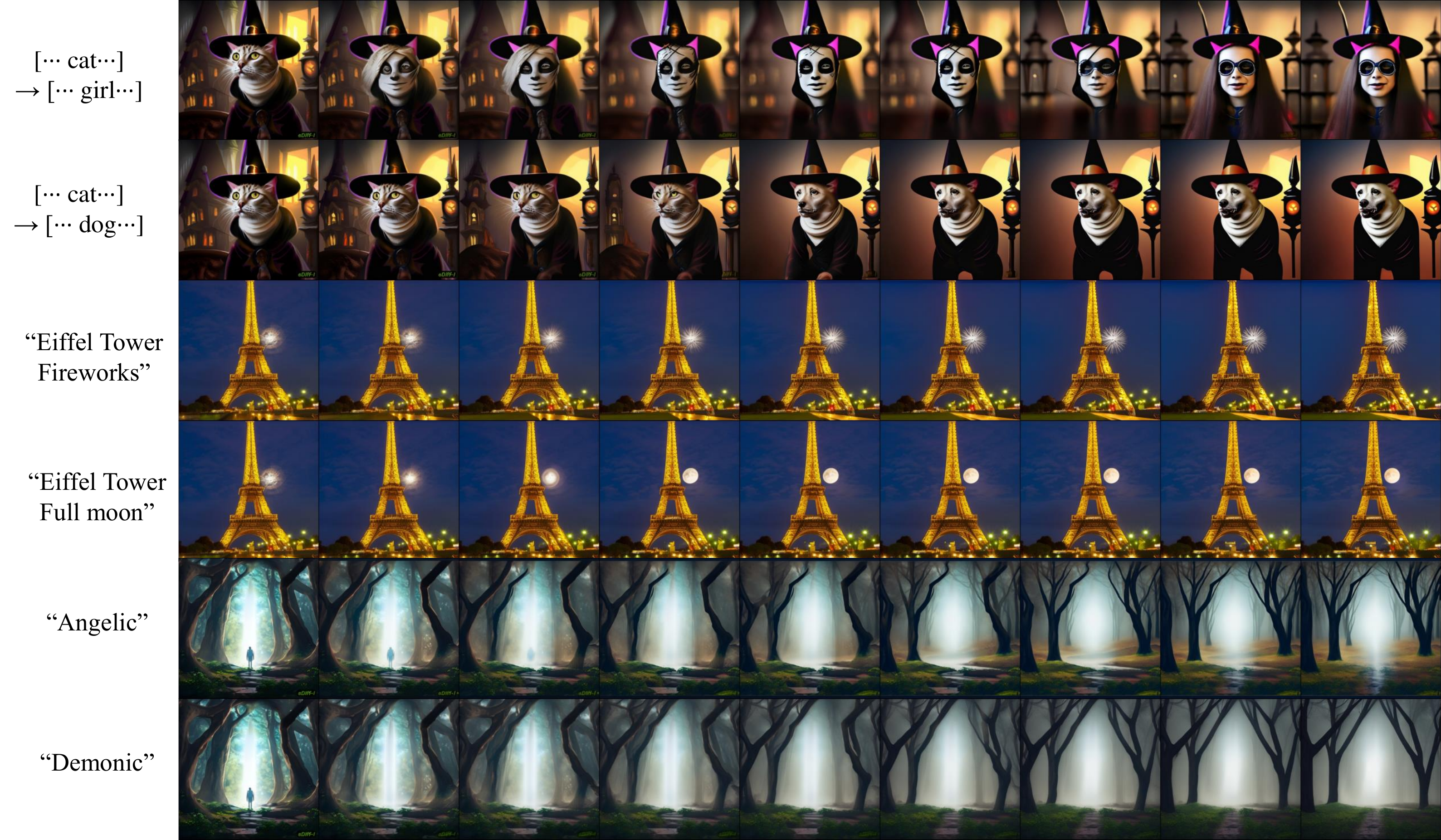}
    \caption{
    \textbf{More examples of image editing using latent basis vectors discovered with various prompts.} }
    \label{fig:stable_text_more}
\end{figure}

\begin{figure}[!h]
    \centering
    \includegraphics[width=1.0\linewidth]{figure/top.pdf}
    \includegraphics[width=1.0\linewidth]{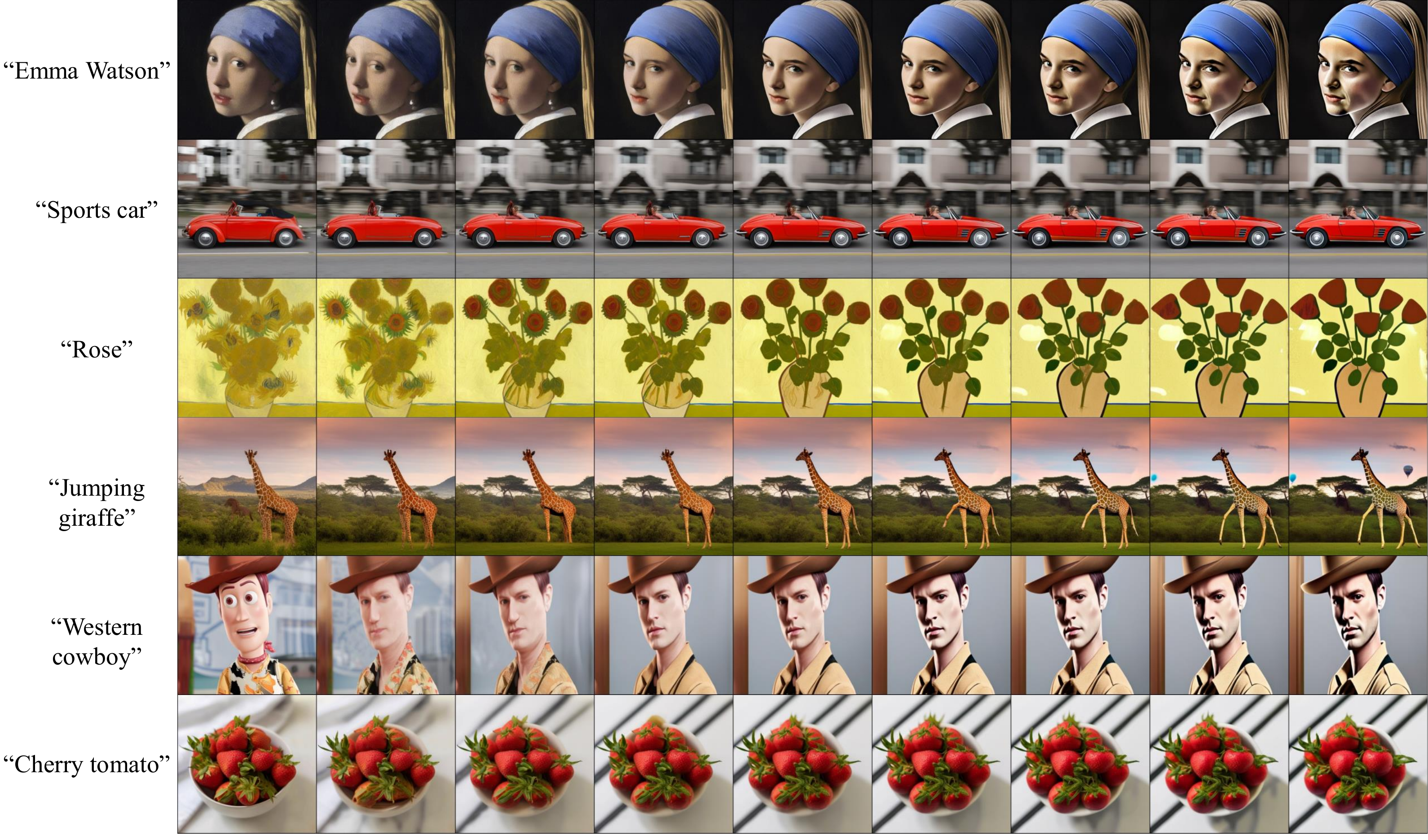}
    \caption{
    \textbf{More examples of image editing using latent basis vectors discovered with various prompts.} }
    \label{fig:stable_text_more2}
\end{figure}

We provide additional examples of image editing using latent basis vectors discovered with various prompts. \fref{fig:stable_text_more}, \ref{fig:stable_text_more2} show image editing with various pictures and various prompts. 
%The prompt of ``[···cat··]'' is ``a cat dressed as a witch wearing a wizard hat in a haunted house''. 
For brevity, we denote the prompt "a cat dressed as a witch wearing a wizard hat in a haunted house" by "[···cat··]" in \fref{fig:stable_text_more}.

\clearpage

% \subsection{More discussion of using latent basis vectors discovered with various prompts}
\subsection{More discussion on the editing capability of the latent basis discovered with text conditions}
\label{appen:more_discussion}

\begin{figure}[!h]
    \centering
    \includegraphics[width=0.8\linewidth]{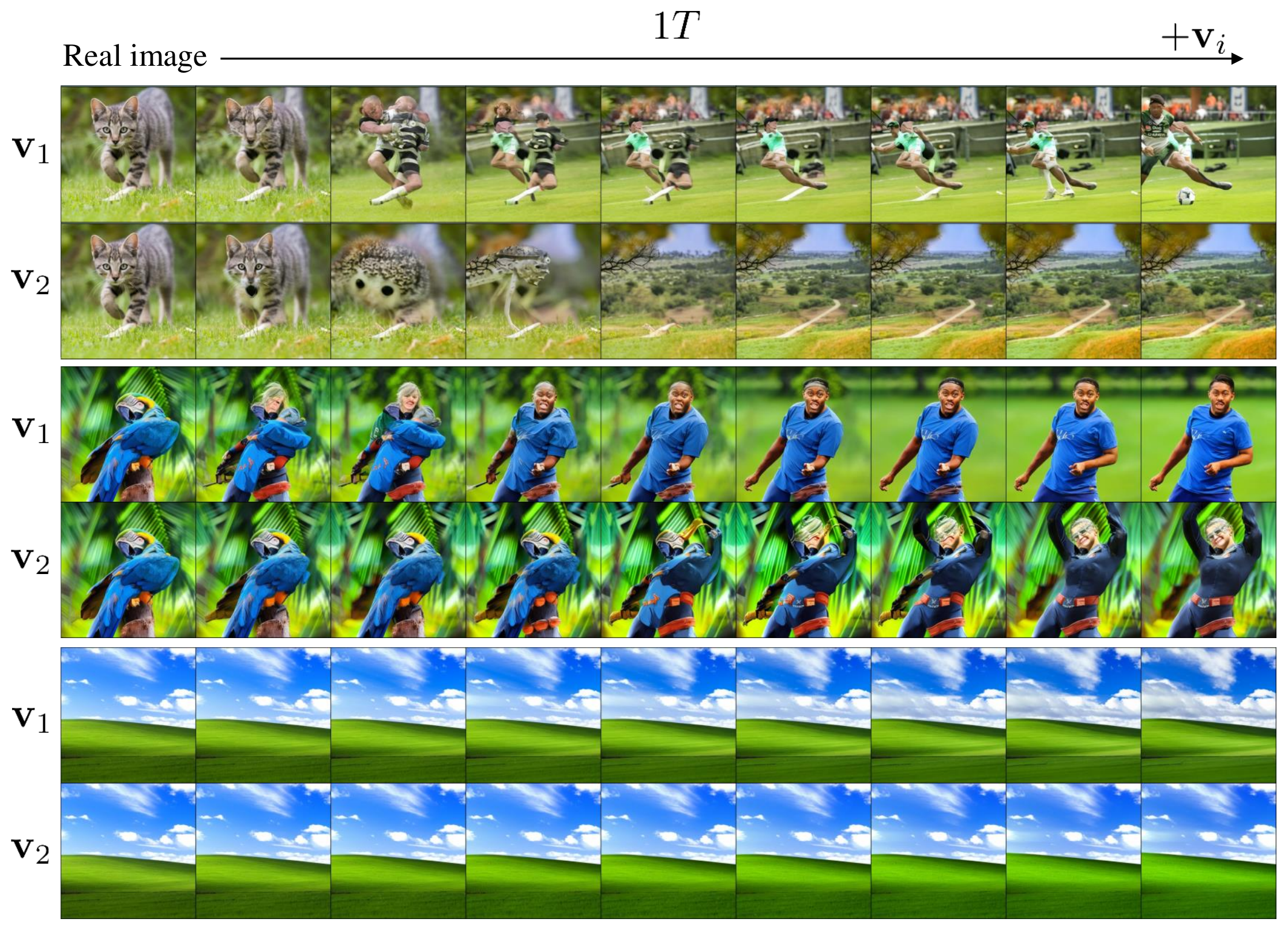}
    \caption{
    \textbf{Failure cases of image editing using the latent basis at 1$T$.} The editing result using $\vv_i$' in Stable diffusion at 1$T$.}
    \label{fig:stable_1T}
\end{figure}

In this subsection, we provide a discussion based on the failure cases of our approach. \fref{fig:stable_1T} shows the results of latent basis found at $t=T$ with Stable diffusion. Unlike unconditional models, the directions found at $t=T$ exhibit rapid and drastic unexpected changes. 
%However, it was observed that for landscape photos without a main object, the expected editing effects were demonstrated at any timesteps. 
However, landscape photos, which do not contain a main object, exhibited desired editing effects at any timesteps.
Moreover, in the case of landscapes, it is conjectured that the latent basis plays a significant role in representing patterns and textures. 
Analyzing the landscape images generated by Stable diffusion would be an interesting topic.

\begin{figure}[!h]
    \centering
    \includegraphics[width=1.0\linewidth]{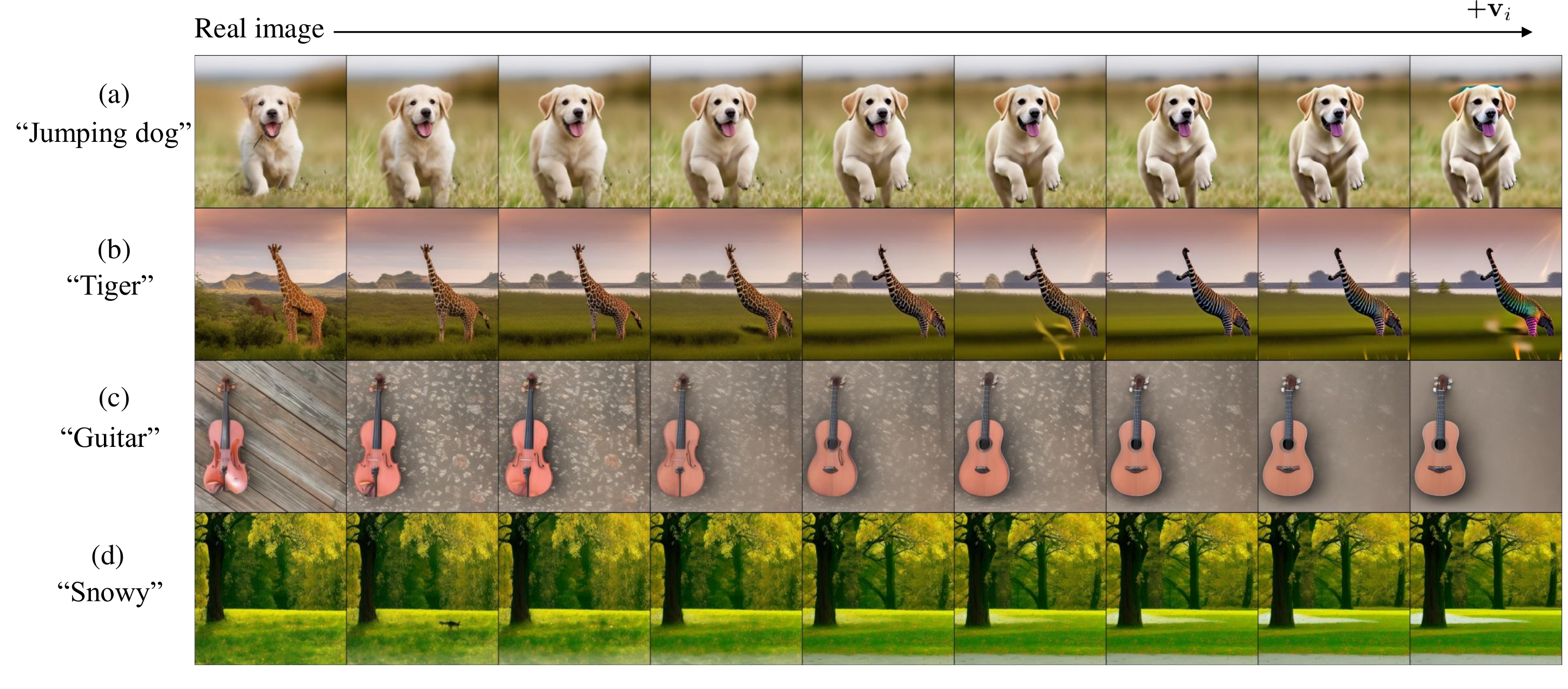}
    \caption{
    \textbf{Failure cases of using prompts.} (a) When using pose or action as a prompt, there are instances where the identity is not preserved.
(b) When the shape of the target subject differs significantly from the source image, the results are often unsatisfactory.
(c) There are cases where the preservation of the background is not achieved.
(d) It is challenging to make significant changes to the entire image.}
    \label{fig:more_limitations}
\end{figure}

\fref{fig:more_limitations} presents examples of failure cases in our image editing using latent basis vectors discovered with various prompts.
(a) When using pose or action as a prompt, there are instances where the identity is not preserved.
(b) When the shape of the target subject differs significantly from the source image, the results are often unsatisfactory.
(c) There are cases where the preservation of the background is not achieved.
(d) It is challenging to make significant changes to the entire image.

Regarding the reasons for these failure cases, we emphasize two factors.
First, we manipulate in the \exspace{}. % ``$\vx_t$''. 
% We adopt a direct manipulation of $\vx_t$ to analyze the latent space of the diffusion model. 
The result in \fref{fig:more_limitations} (a) implies that \exspace{} is not a space where disentanglement for identity is achieved effectively.
On the other hand, in \ehspace{}, there are results indicating successful preservation of identity \cite{kwon2022diffusion, haas2023discovering}. 
Investigating the disentanglement capability of \exspace{} and any other distinguishing features it may have compared to \ehspace{} would be an interesting future research topic.

Secondly, we perform manipulation by adding and subtracting the "signal" that the model pays attention to in $\vx_t$.
% The directions we find represent the signal captured by the model's features. 
Here, The signal is captured from the current input $\vx_t$, which limits the deviation from the original form. Therefore, when there is a substantial difference in shape, such as transforming a giraffe into a tiger, the results may not be satisfactory. (\fref{fig:more_limitations} (b))
When we utilize text conditions, the latent basis aligns with the text information. This leads to not capturing background information, resulting in changes in the background when manipulated. It is also an interesting research topic to capture signals related to the background. (\fref{fig:more_limitations} (c))
% Since we utilize the top $n$ signals captured by the features, they often do not include background information. It is also an interesting research topic to capture signals related to the background. (\fref{fig:more_limitations} (c))
Since the model focuses on finer features as t approaches 0, if broad changes are desired, manipulation should be performed at $t=T$. However, manipulation at $t=T$ is unstable. Deep analysis of $\vx_t$ at $t=T$ in Stable diffusion is also an intriguing research topic. (\fref{fig:more_limitations} (d))

% Regarding the reasons for these failure cases, we emphasize two factors.
% First, the ``signal''. The directions we find represent the signal captured by the model's features. This signal is entirely determined by the given text, which can result in a loss of identity preservation. The signal is captured from the current input $\vx_t$, which limits the deviation from the original form. Therefore, when there is a substantial difference in shape, such as transforming a giraffe into a tiger, the results may not be satisfactory. Since we utilize the top $n$ signals captured by the features, they often do not include background information. It is also an interesting research topic to capture signals related to the background.

Despite these limitations, we have successfully achieved direct manipulation in the latent space $\vx_t$ at a single timestep, which, to our knowledge, is the first of its kind. Through this, we provide insights into the model and contribute to the understanding of the latent space, hopefully benefiting the diffusion community.

